\pdfoutput=1
\documentclass[10pt,twocolumn]{IEEEtran}

\usepackage{cite}
\usepackage{amsmath,amssymb,amsfonts}
\usepackage{graphicx}
\usepackage{textcomp}
\usepackage{algorithmic}
\usepackage{array}
\usepackage{tabularx}
\usepackage{multirow}
\usepackage{booktabs}
\usepackage{hhline}
\usepackage{bm}
\usepackage{url}
\usepackage{stfloats}
\usepackage{placeins}
\usepackage{multicol}

\usepackage{capt-of}

\usepackage{cuted}
\usepackage{subcaption}

\graphicspath{{figures/}}

\newcolumntype{Y}{>{\raggedright\arraybackslash}X}
\newcolumntype{P}[1]{>{\raggedright\arraybackslash}p{#1}}
\newcolumntype{C}[1]{>{\centering\arraybackslash}p{#1}}

\usepackage[font=footnotesize,labelfont=bf,labelsep=period]{caption}

\title{CrowdH-Edit: Reference-Guided Instruction-Based Synthetic Hajj Crowd Image Generation for Crowd Counting}

\author{
\begin{minipage}{0.98\textwidth}
\centering
Amirah F. Alshammari$^{1,2}$, Bander A. Alzahrani$^{3}$, and Nahed A. Alowidi$^{4}$\\[0.8em]
\small
$^{1}$Information Systems Department, Faculty of Computing and Information Technology, King Abdulaziz University, Jeddah, Saudi Arabia\\
$^{2}$Department of Information Systems, College of Computer and Information Sciences, Jouf University, Aljouf, Saudi Arabia\\
$^{3}$Information Systems Department, Faculty of Computing and Information Technology, King Abdulaziz University, Jeddah, Saudi Arabia\\
$^{4}$Computer Science Department, Faculty of Computing and Information Technology, King Abdulaziz University, Jeddah, Saudi Arabia\\[0.5em]
E-mails: aalshammari0114@stu.kau.edu.sa, baalzahrani@kau.edu.sa, nalowidi@kau.edu.sa
\end{minipage}
}

\date{}

\begin{document}

\renewcommand{\thetable}{\arabic{table}}
\renewcommand{\thefigure}{\arabic{figure}}

\maketitle

\begin{abstract}
The scarcity of domain-specific annotated images remains a major limitation for developing computer-vision systems for Hajj crowd analysis. Existing crowd-counting datasets and unconstrained generative models do not capture Hajj-specific visual conditions, including sacred-site geometry, white-clothed pilgrim crowds, and indoor/outdoor illumination variation. This study presents CrowdH-Edit, a reference-guided, instruction-based framework for synthetic Hajj crowd-image generation. Real Hajj images provide structural references, while textual instructions guide changes in crowd conditions and illumination and help preserve the main scene layout. Stable Diffusion 2.1 and Stable Diffusion XL 1.0 with low-rank adaptation were evaluated as comparative baselines. Qwen Image Edit was used without additional training within a ComfyUI workflow and selected for final dataset construction. The workflow produced  CrowdH-Edit, a cleaned pool of 6,122 synthetic images, from which 387 images were selected and annotated to construct CrowdH-Edit-Ann-387. Among the evaluated generation pipelines, Qwen Image Edit achieved the highest semantic-alignment score and the lowest distributional and perceptual distances. Downstream evaluation across five crowd-counting models showed that adding the annotated synthetic subset to real training data reduced mean absolute error for three models and root mean squared error for two. These findings indicate that reference-guided synthetic Hajj imagery can support crowd-counting training, although its benefits depend on the model and evaluation metric.
\end{abstract}

\vspace{0.2em}

\section{Introduction}

Crowd counting is a fundamental task in crowd analysis and supports safety monitoring, emergency planning, and operational decision-making in dense public environments. It is particularly important during the Hajj event, where crowd density, viewpoint, illumination, occlusion, and movement patterns can change rapidly across different ritual locations. In such large-scale pilgrimage environments, reliable crowd-counting systems require not only robust models but also representative training data that capture the visual complexity of Hajj scenes.

Recent crowd-counting research has achieved significant contributions through improved network architectures for handling scale variation \cite{b1,b2,b3,b4,b5}, extracting multi-resolution features \cite{b6,b7}, and developing more effective loss functions and supervision strategies  \cite{b8,b9}. These advances have significantly improved the accuracy of counting models on standard benchmarks. However, these models remain strongly dependent on the quality, diversity, and annotation reliability of the training data. When the training distribution does not match the target environment, model performance may degrade due to domain shift. This issue is particularly relevant to Hajj crowd analysis, where existing general-purpose datasets do not fully represent sacred-site geometry, white-clothed pilgrim crowds, ritual-specific spatial organization, indoor/outdoor illumination variation, and severe occlusion in highly congested scenes.

Existing real crowd-counting datasets have contributed substantially to the development of the field, but they do not fully satisfy the requirements of Hajj-specific crowd analysis. Real datasets are often limited in scale, scene diversity, viewpoint coverage, or annotation consistency \cite{b6,b10,b11,b12,b13}. Constructing real Hajj crowd datasets is challenging because image collection and annotation are expensive, time-consuming, and constrained by privacy, accessibility, and cultural-sensitivity considerations. Dense head-point annotation is particularly difficult in congested scenes because heads may be small, partially occluded, or visually ambiguous \cite{b14,b15}.

Synthetic data provides an alternative way to increase dataset size and expand visual coverage. Simulation-based datasets such as GCC \cite{b16} and CrowdX \cite{b17} demonstrate that synthetic crowd data can support crowd-counting training and enable controlled analysis of factors such as camera viewpoint, background, crowd density, and environmental conditions \cite{b17}. Nevertheless, simulation-based datasets often exhibit a synthetic-to-real appearance gap and are mainly designed around urban or game-engine environments. They may not reproduce the spatial structure, and architectural characteristics of Hajj scenes. Therefore, there is a need for synthetic data-generation approaches that can expand Hajj crowd imagery while preserving real-scene structure.

Recent works have used pretrained text-to-image diffusion models for image editing \cite{b18,b19,b20,b21}. However, text-to-image diffusion models remain difficult to use for targeted editing because small prompt changes may produce unpredictable variations in scene layout, object placement, and visual structure \cite{b22}. To obtain localized edits, many studies fall back to inpainting, which requires a user-provided spatial mask and careful boundary handling. This adds manual effort and limits scalability for large dataset construction \cite{b18}. These limitations motivate instruction-based image-editing methods \cite{b23,b24}, where the input consists of a real reference image and an action-oriented textual instruction describing the intended modification. This formulation enables more controlled editing while retaining most of the original scene structure.

Recent instruction-capable editors' approach, such as Qwen Image Edit \cite{b24}, make this paradigm relevant for geometry-sensitive dataset construction. Qwen Image Edit allows users to guide the editing process through natural-language instructions, such as modifying crowd level, illumination, or scene conditions, while using the input image as a structural reference. This capability is critical for Hajj crowd images because sacred-site geometry, architectural boundaries, and crowd organization must remain visually consistent. Furthermore, Qwen Image Edit can be implemented through a transparent node-based ComfyUI workflow \cite{b25}, which substantially simplifies the entire generation process and makes the model easy to deploy, test, and iterate on.The visual node-based interface reduces the need for custom scripting and facilitates modular workflow configuration, allowing the model to be set up and experimented with quickly through modular, reusable components. This workflow also supports reproducible execution and systematic generation of edited outputs, so new editing scenarios can be tested rapidly by adjusting individual nodes rather than rebuilding the pipeline. Thus, instruction-guided reference editing is delivered through an accessible and efficient ComfyUI workflow. It provides a practical mechanism for producing controlled synthetic Hajj crowd images without requiring user-drawn masks or full output-image descriptions.

Therefore, in this paper, we introduce CrowdH-Edit, a dataset of 6,122 synthetic Hajj crowd images generated using Qwen Image Edit within a reference-guided ComfyUI workflow. The dataset was obtained by applying automated quality checks and human visual screening to an initial pool of 8,000 Qwen-generated candidate images. From the cleaned dataset, 387 images were selected and annotated with head points to construct CrowdH-Edit-Ann-387 for downstream crowd-counting evaluation. The proposed datasets are designed to mitigate data scarcity in Hajj crowd-counting scenarios by generating controlled visual variations from real reference scenes while preserving key architectural and spatial structures. The generation process focuses on modifying crowd level and illumination conditions across representative pilgrimage locations, including Tawaf/Kaaba, Mina, Arafat, Al-Jamarat, and Safa-Marwah. The main contributions of this paper are summarized as follows:

\begin{itemize}

\item We design a reference-guided instruction strategy tailored to Hajj crowd imagery and integrate it into a Qwen Image Edit--ComfyUI workflow. The instruction design controls crowd-level conditions, illumination, reference-view consistency, and location-specific preservation constraints while using real Hajj images as structural references to help preserve scene layout and location-specific structures.

\item We construct CrowdH-Edit, a cleaned dataset of 6,122 synthetic Hajj crowd images retained from an initial pool of 8,000 Qwen-generated candidates, and CrowdH-Edit-Ann-387, a 387-image subset with head-point annotations for supervised crowd counting validation.

\item We evaluate the proposed dataset-construction approach at two levels. First, Qwen Image Edit is compared with Hajj-adapted
SD 2.1 + LoRA and SDXL 1.0 + LoRA baselines under a controlled evaluation protocol using complementary distributional, structural, perceptual, and semantic metrics. Second, the practical utility of CrowdH-Edit-Ann-387 is assessed across five crowd-counting models under real-only and real-plus-synthetic training settings using a fixed real held-out test set.

\end{itemize}

The remainder of this paper is organized as follows. Section II reviews related work on crowd-counting methods, real and synthetic crowd datasets, and generative image-editing approaches. Section III presents the proposed methodology, including reference data preparation, Qwen Image Edit generation, dataset statistics, instruction design, data cleaning, annotation, and evaluation protocol. Section IV reports and discusses the experimental results, including image-quality assessment and crowd count validation. Section V discusses limitations. Finally, Section VI concludes the paper and outlines future research directions.

\section{Related Work}

\subsection{Crowd Counting Methods}
Crowd-counting research has evolved from traditional hand-crafted approaches to deep learning-based density estimation and localization models. Traditional methods include detection-based \cite{b26,b27}, regression-based \cite{b12,b28}, and density-estimation approaches \cite{b29,b30}. Detection-based methods aim to localize people or body parts before counting, but their performance often degrades in dense scenes because of occlusion, scale variation, and complex backgrounds. Regression-based methods directly learn a mapping from image features to crowd counts, while density-estimation methods predict spatial density maps that better preserve crowd distribution information.

Deep learning-based approaches have become dominant because of their stronger feature representation capability. CNN-based methods \cite{b6,b31,b32} extract local and multi-scale features for density-map estimation, but their limited receptive fields can restrict global context modeling. ViT-based methods \cite{b33,b34}use self-attention to capture long-range contextual information, while joint CNN–ViT methods \cite{b35,b36} combine local feature extraction with global dependency modeling. More recent approaches have explored diffusion model-based methods \cite{b37,b38} formulate density-map generation as a denoising process, and multimodal methods \cite{b39,b40} integrate additional modalities such as thermal or depth information to improve robustness. These methods have improved counting accuracy on standard benchmarks, but their performance remains dependent on representative and accurately annotated training data.

\subsection{Real, Synthetic, and Hajj-Specific Crowd Datasets}
Crowd-counting datasets play a central role in training and evaluating counting models because dataset scale, scene diversity, annotation quality, and viewpoint variation strongly affect model generalization. Therefore, real publicly available crowd-counting datasets have evolved in scale, scene diversity, resolution, and annotation richness. Early datasets such as UCSD \cite{b12} and Mall \cite{b28} were collected from fixed surveillance cameras and remain useful for controlled monitoring tasks. But their limited density range, low resolution, and restricted viewpoints reduce their suitability for highly congested crowd analysis. UCF-CC-50 \cite{b11} introduced more extreme density variation across a small number of images, while WorldExpo’10 \cite{b13} provided broader scene coverage from surveillance-style video frames. ShanghaiTech \cite{b6} became one of the most widely used benchmarks, with Part A containing dense internet-sourced scenes and Part B covering relatively sparse street scenes. More recent large-scale datasets including UCF-QNRF \cite{b41}, JHU-Crowd++ \cite{b42}, and NWPU-Crowd \cite{b43}, increased image resolution, annotation volume, density variation, and environmental diversity, providing more challenging benchmarks for deep crowd-counting models. However, existing real datasets are important but general-purpose.These general-purpose benchmarks were not designed specifically to represent Hajj environments. As a result, models trained only on these benchmarks may experience domain-shift when applied to pilgrimage scenes with different viewpoints, illumination conditions, crowd organization, and occlusion patterns.

Synthetic and generative data have been proposed as an alternative strategy to reduce annotation cost and increase data diversity. Simulation-based datasets generate crowd scenes using game engines or virtual environments, which enables automatic labels and controllable variations in crowd density, viewpoint, illumination, and camera configuration. GCC introduced a large-scale synthetic crowd-counting dataset collected from the Grand Theft Auto V environment and used it for synthetic-to-real crowd-counting transfer \cite{b16}. In addition, CrowdX explored parameterized synthetic crowd generation with accurate labels and controllable variations to improve crowd-counting performance \cite{b17}, while CVCS extended this direction to cross-view and cross-scene multi-view crowd counting by generating synthetic scenes with multiple camera views \cite{b44}. Recently diffusion-based data augmentation methods have also been explored for counting tasks. For example, diffusion-based object-counting augmentation conditioned on dot maps and smoothed density maps has shown that synthetic images can support counting-model training on standard benchmarks \cite{b38}. However, simulation-based datasets and general diffusion-based augmentation methods are typically designed around general crowd scenes, urban environments, or benchmark datasets rather than Hajj-specific pilgrimage scenes.

Several Hajj-related datasets and resources have been developed for different purposes. Existing Hajj resources include datasets for event recognition, behavior analysis, density classification, surveillance-based crowd monitoring, and multimodal sensing \cite{b45,b46,b47,b48,b49,b50,b51}. However, these datasets remain limited in scale, accessibility, scene coverage, and annotation. Some Hajj-specific image datasets focus on limited locations such as Tawaf or provide restricted accessibility, while others emphasize behavior analysis or density categories rather than dense head-point annotations suitable for training crowd-counting models.

Overall, existing crowd-counting datasets provide valuable benchmarks for model development and evaluation, but do not explicitly capture the visual and spatial characteristics of Hajj environments. They lack representation of sacred-site geometry, white-clothed pilgrim crowds, ritual-specific spatial organization, and the strong illumination differences between indoor and outdoor pilgrimage scenes. Although several Hajj-related datasets have been introduced, they remain limited in accessibility, scene coverage, annotation format, or suitability for supervised crowd-counting training. These limitations highlight the need for domain-specific Hajj crowd datasets that better reflect the visual conditions of pilgrimage environments.

\subsection{Generative and Instruction-Guided Editing Approaches for Synthetic Image Data}

Conventional augmentation strategies, such as flipping, cropping, color jittering, and scaling can increase sample size. However, the augmented images often remain highly correlated with the original data which may not provide sufficient domain diversity. Therefore, generative models have been increasingly explored for synthetic data construction because they can expand visual diversity and reduce the cost of manual data collection and annotation. Generative models provide an alternative strategy for expanding training data by learning visual patterns from real images and producing synthetic samples. Early conditional image-to-image translation methods and GAN-based models demonstrated that realistic images can be synthesized or transformed across domains using paired or unpaired adversarial learning \cite{b52,b53} Conditional image synthesis has also been investigated specifically for Hajj crowd scenes \cite{b57}. However, GAN-based models can be difficult to optimize and may suffer from mode collapse, which limits their ability to represent the full diversity of complex crowd scenes. 

Diffusion models have recently achieved strong image-generation performance through iterative denoising processes \cite{b54}. Latent diffusion models improved the practicality of high-resolution synthesis by operating in a compressed latent space \cite{b55}, and large-scale diffusion variants such as SDXL further improved visual fidelity and resolution \cite{b56}. These developments make diffusion-based generation attractive for synthetic image dataset construction. Nevertheless, unconstrained text-to-image generation is not always suitable for geometry-sensitive crowd scenes. In structured Hajj environments, small prompt changes may alter landmark geometry, crowd placement, object structure, or scene layout. Such instability is problematic for crowd-counting research because the generated data should preserve realistic spatial organization, architectural context, and crowd appearance.

To improve controllability, recent work has shifted from unconstrained generation toward image editing. SDEdit performs guided editing by perturbing a reference image and denoising it under a new condition \cite{b20}, while Prompt-to-Prompt improves prompt-based editing consistency by controlling cross-attention during modification \cite{b22}. Instruction-based image editing further simplifies the process by allowing the model to modify an input image according to a textual command rather than requiring a complete target-image description \cite{b23}. This direction is particularly relevant for domain-specific dataset construction because the input image can provide structural priors while the instruction controls the intended visual variation.

Qwen-Image extends this direction by supporting both image generation and precise image editing \cite{b24}. It is a multi-task training paradigm that integrates text-to-image, text-image-to-image, and image-to-image reconstruction tasks. It also uses semantic representations from Qwen2.5-VL and reconstructive representations from a VAE, which helps balance semantic coherence with visual fidelity during editing \cite{b24}. These properties make Qwen Image Edit relevant for reference-guided synthetic dataset construction, when the target application requires controlled modification rather than unconstrained scene generation.

The reviewed literature shows that crowd-counting models have advanced substantially, and both real and synthetic datasets have supported broader benchmarking and training. However, existing crowd-counting datasets do not adequately capture Hajj-specific visual conditions, including sacred-site geometry, white-clothed pilgrim crowds, ritual-specific spatial organization, and rapidly changing indoor/outdoor illumination. In addition, simulation-based datasets provide scale and automatic labels but may retain a synthetic-to-real appearance gap, whereas unconstrained text-to-image generation can alter geometry and hallucinate scene elements. Existing Hajj-related datasets are useful but remain limited in accessibility, coverage, annotation format, or suitability for supervised image-based crowd counting. These limitations motivate a reference-guided instruction-based editing strategy that uses real Hajj images as structural priors and applies controlled textual instructions to generate domain-specific synthetic data for downstream crowd-counting validation.

\section{Methodology}
\subsection{Overview of the CrowdH-Edit Dataset Construction Framework}
\begin{figure}[!htbp]
\centering
\includegraphics[width=\columnwidth, height=5cm]{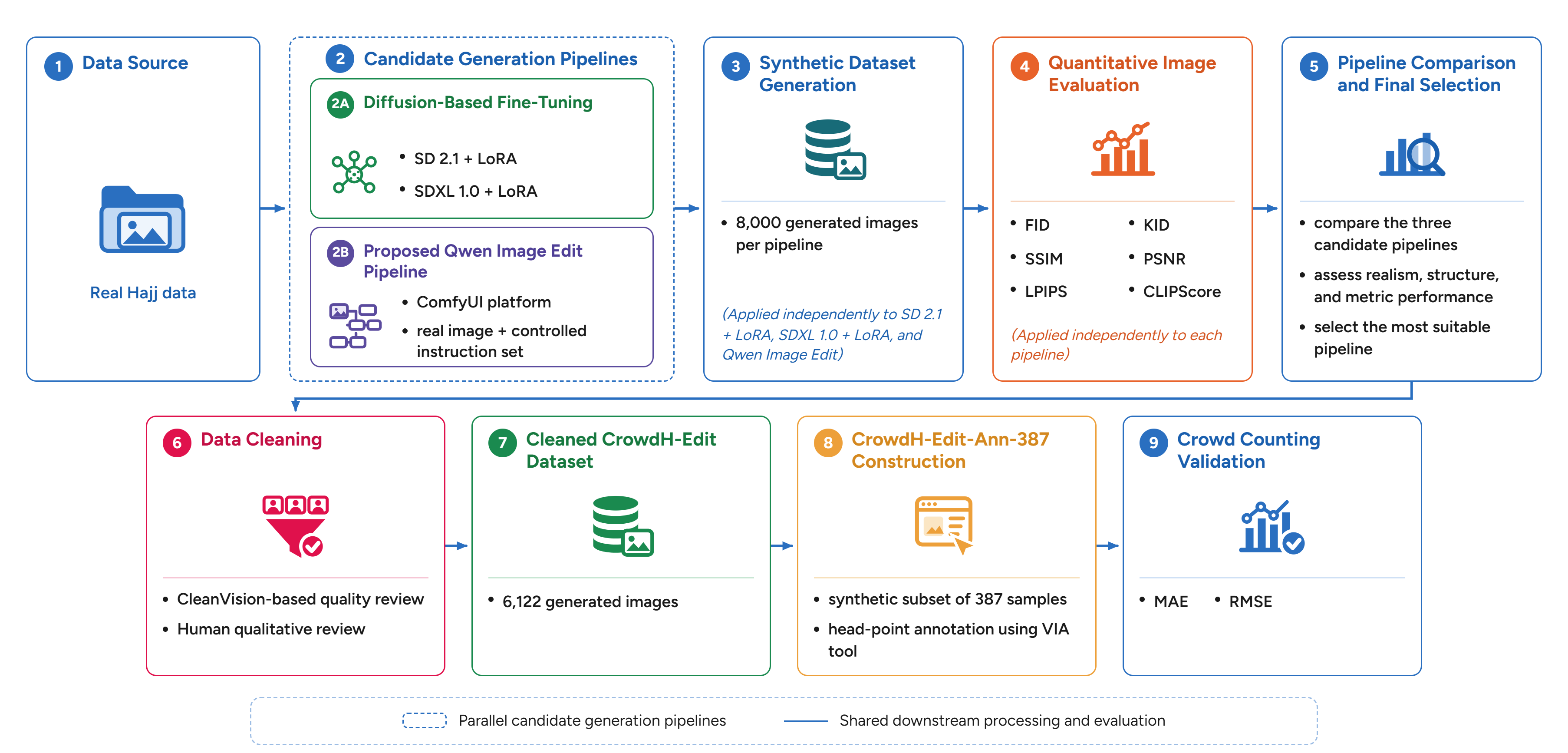}
\caption{Overall methodology used to construct and validate the proposed CrowdH-Edit dataset.}
\label{Fig_1}
\end{figure}

The proposed CrowdH-Edit framework constructs a synthetic Hajj crowd dataset for crowd-counting research. As illustrated in Figure~\ref{Fig_1}, it integrates reference-guided generation, pipeline comparison, quality screening, and annotation. Real images from five Hajj locations \cite{b57} provide structural references, while textual conditions guide changes in crowded conditions and illumination.

Three generation pipelines were investigated within ComfyUI. SD 2.1 + LoRA and SDXL 1.0 + LoRA served as fine-tuned diffusion baselines \cite{b58,b59,b60}. Qwen Image Edit was used as a pretrained instruction-based editor without additional fine-tuning \cite{b61,b24}. Each pipeline produced 8,000 candidate images. Quantitative evaluation and qualitative inspection assessed image quality, instruction alignment,and reference-scene preservation. Qwen Image Edit was selected for dataset construction based on this comparison.

The selected image pool underwent CleanVision-based quality review and human visual screening, yielding 6,122 images in the cleaned CrowdH-Edit dataset. From this pool, 387 images were selected for their visual quality, clear head regions, and coverage of different Hajj locations. Head-point annotations were created using VIA to construct CrowdH-Edit-Ann-387. Its usefulness for crowd counting was then assessed by comparing real-only training with training augmented by this annotated synthetic subset.

\FloatBarrier

\subsection{Reference Data Preparation}

The generative-pipeline development retained the 792-frame training set and 201-frame validation set used in the previous study \cite{b57}. These images covered five Hajj locations: Kaaba/Tawaf, Mina, Arafat, Al-Jamarat, and Safa-Marwah. An additional independent set of 261 real Hajj frames, not included in the training or validation sets, was reserved as generative test set. Source-video IDs were checked to ensure that no source video represented in the 261-frame controlled evaluation set appeared in the training or validation sets. The separate 77-image real test set used for downstream crowd-counting evaluation was not used in the generative-model training, workflow-setting selection, or controlled generation evaluation.

To prevent source video overlap, the 261-frame generative evaluation set and the separate 77-image downstream test set were drawn from source videos with IDs distinct from those used in the training and validation data; the 77-image test set was also video-disjoint from the 261-frame generative
evaluation set. Further details of the data collection and preprocessing protocol are provided in~\cite{b57}.

The 792-frame training set was used to fine-tune the LoRA adapters for SD 2.1 and SDXL 1.0, whereas Qwen Image Edit was used without additional fine-tuning. The 201-frame validation set was used to assess and select the model and workflow settings across the three generation pipelines. After these settings were fixed, the same 261 held-out frames served as common reference inputs for the controlled comparison of SD 2.1 + LoRA, SDXL 1.0 + LoRA, and Qwen Image Edit. 

\subsection{Candidate Generation Pipelines}

\subsubsection{Diffusion-Based Fine-Tuning Baselines}

Two diffusion-based baselines were implemented to evaluate whether parameter-efficient adaptation could generate realistic and structurally consistent crowd scenes before adopting the final reference-guided editing workflow. The evaluated baselines were SD 2.1 + LoRA and SDXL 1.0 + LoRA. Both used Low-Rank Adaptation (LoRA) to fine-tune adapter weights on the same 792 real Hajj reference images. The base models were not trained from scratch. This approach reduced training costs and supported adaptation to pilgrimage-site appearance, white-clothed crowds, and illumination variation.

To prepare textual conditioning for the diffusion-based baselines, a structured captioning strategy was adopted. First, CLIP-based annotation was used to extract key visual metadata from the image name, including location, time-of-day condition, and crowd-density level. Then, BLIP~\cite{b62} was employed to generate an initial image-level caption, which was further enriched with Hajj-specific contextual terms. This strategy combined image-level captions with Hajj-specific contextual information, making it suitable for guiding the Stable Diffusion baselines in generating Hajj crowd scenes with improved contextual consistency.

The SD 2.1 + LoRA baseline was selected for the initial prototyping stage because of its lightweight architecture, moderate graphics-memory requirements, and well-established LoRA fine-tuning infrastructure. It used the v21\_768ema-pruned.safetensors checkpoint and was fine-tuned at 768 × 768 resolution, which is consistent with the native 768-pixel Stable Diffusion 2.1 checkpoint and supports faster experimentation under moderate hardware constraints. The LoRA adapter used rank 16, LoRA alpha 32, AdamW 8-bit optimization, fp16 mixed precision. The micro-batch size was 1, with gradient accumulation of 4, and an effective batch size of 4. The UNet learning rate was $1 \times 10^{-4}$, while the text encoder was kept frozen to preserve the pretrained language representation. Training was conducted for 3 epochs using latent caching, gradient checkpointing. The maximum gradient norm was 1.0, and a warm-up ratio was 0.1. Flip and color augmentations were disabled to avoid unrealistic transformations of sacred-site orientation, Kaaba/Tawaf geometry, and Hajj-specific color appearance. Aspect-ratio bucketing was enabled to support mixed image aspect ratios while training at the target resolution. This baseline was computationally efficient and useful for feasibility testing, but the lower resolution limited fine architectural detail preservation.

The SDXL 1.0 + LoRA baseline was included to examine whether higher model capacity and resolution could improve architectural detail, crowd texture, and scene fidelity. It used the sdXL\_v10VAEFix.safetensors checkpoint with the SDXL model type flag and the standard LoRA network module. It was fine-tuned  under the structured captioning strategy described above. The adapter used LoRA rank 16, LoRA alpha 1, AdamW 8-bit optimization, and bfloat16 mixed precision. The micro-batch size was 1, with gradient accumulation of 4 and an effective batch size of 4. The UNet learning rate was set to $1 \times 10^{-4}$, and both SDXL text encoders were trained with a learning rate of  $1 \times 10^{-4}$, allowing the textual conditioning to better associate Hajj-specific caption with visual content. Training was conducted for 8 epochs at 1024~$\times$~1024 resolution. Gradient checkpointing was enabled to reduce memory usage, and the multi-resolution noise discount was set to 0.3 to support fine-detail learning. Aspect-ratio bucketing was disabled to enforce square crops and reduce geometric distortion in structured Hajj scenes. 

\subsubsection{Proposed Qwen Image Edit Pipeline}

The proposed generation pipeline uses Qwen Image Edit as an inference-time, reference-guided image editor within a ComfyUI workflow. In contrast to the diffusion-based LoRA baselines, Qwen Image Edit was not fine-tuned in this study. Instead, 261 real Hajj images were used as structural references, and controlled textual instructions were used to guide the editing process. This design enables targeted modification of selected visual attributes, such as crowd-level condition and illumination, while helping preserve the spatial structure of the original Hajj scene. This property is particularly important for Hajj image, where sacred-site geometry, architectural layout, and crowd organization, people head appearance in the crowd must remain visually consistent for crowd-counting research. The editing instructions are described in Subsection 3.4.

\begin{figure*}[!htbp]
\centering
\includegraphics[width=\textwidth,height=8cm,
keepaspectratio]{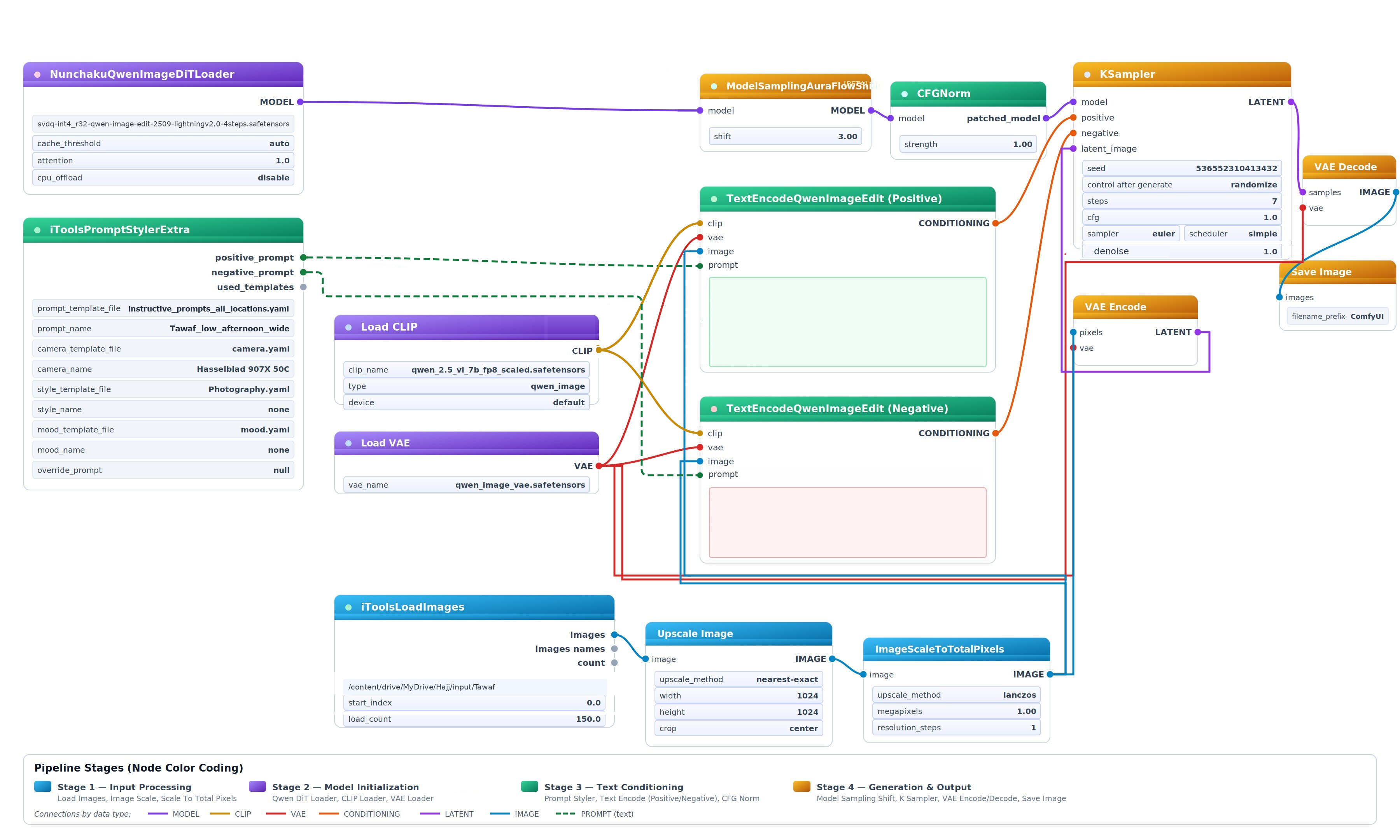}
\caption{Qwen Image Edit workflow implemented in ComfyUI.}
\label{fig:Fig_2}
\end{figure*}

Qwen Image Edit was included as a candidate because its reference-guided editing design combines text-based instruction conditioning through Qwen2.5-VL with reference-image conditioning and latent image encoding/decoding through the Qwen Image VAE. Compared with unconstrained text-to-image generation, this design provides several advantages for Hajj crowd-scene synthesis. First, the reference image acts as a structural prior, helping to preserve scene geometry, architectural layout, and location-specific visual elements. Second, editing from real Hajj images supports more plausible crowd appearance and spatial organization. Third, instruction-based editing enables targeted modifications, such as changes in crowd level or illumination, while using the reference image to condition the generation process. Finally, using Qwen Image Edit as an inference-time editor reduces the need for extensive location-specific fine-tuning while supporting controlled synthetic data generation. Based on these advantages, the proposed workflow was implemented in ComfyUI, as shown in Figure~\ref{fig:Fig_2}.

The Qwen Image Edit workflow was implemented using a node-based ComfyUI pipeline. The workflow consists of four main stages, summarized in Table~\ref{tab:qwen_workflow}. First, real Hajj reference images are loaded in batches, resized to 1024~$\times$~1024, and scaled to a standardized pixel budget. Second, the required pretrained components are initialized, including the Qwen Image Edit diffusion transformer (DiT), Qwen text encoder, and VAE. The workflow uses the svdq-int4\_r32-qwen-image-edit-2509-lightningv2.0-4steps.safetensors checkpoint \cite{b61}. No additional fine-tuning of Qwen Image Edit was performed on the Hajj data. Third, positive and negative textual instructions are encoded to guide the desired edits and constrain unwanted changes. The positive instruction specifies the required crowd, illumination, and viewpoint conditions, whereas the negative instruction is used to reduce undesirable artifacts and structural distortions. In the illustrated workflow, the positive and negative instructions are loaded from the selected template entry, identified by prompt\_name, in instructive\_prompts\_all\_locations.yaml and passed to the corresponding text-encoding nodes through their prompt inputs. Accordingly, the text boxes inside the encoding nodes are not used for manual prompt entry in this configuration.Finally, the reference images are encoded into latent space, the edited latent representations are sampled using KSampler, and the outputs are decoded through the VAE to produce the final synthetic images.

%-----------------------------

\begin{table*}[!htbp]
\caption{Qwen Image Edit Workflow Stages Implemented in ComfyUI}
\label{tab:qwen_workflow}
\centering

\renewcommand{\arraystretch}{1.15}
\setlength{\tabcolsep}{5pt}

\begin{tabularx}{\textwidth}{
>{\raggedright\arraybackslash}p{0.18\textwidth}
>{\raggedright\arraybackslash}X
>{\raggedright\arraybackslash}p{0.29\textwidth}
}
\toprule
\textbf{Stage} & \textbf{Main operations} & \textbf{Purpose} \\
\midrule

Input preparation
&
Load reference image, resize to \(1024 \times 1024\), and scale to 1.00 MP
&
Prepare the real Hajj images as structural references for editing
\\

Model initialization
&
Load the Qwen Image Edit DiT checkpoint through the Nunchaku loader, together with the Qwen2.5-VL text encoder and Qwen VAE.
&
Initialize pretrained components required for instruction-based editing
\\

Instruction conditioning
&
Load the instruction template, encode positive and negative instructions, and apply guidance normalization.
&
Specify the intended edit while constraining unwanted visual and structural changes
\\

Generation and output
&
Encode the reference images into latent space, sample the edited latent representations using KSampler, decode them through the VAE, and save the generated images.
&
Generate the final edited synthetic Hajj image
\\

\bottomrule
\end{tabularx}
\end{table*}

The inference configuration used for the Qwen Image Edit pipeline is summarized in Table~\ref{tab:qwen_generation_parameters}. These parameters were selected to balance instruction adherence, visual quality, and computational efficiency. The workflow used Euler sampling, 7 sampling steps, CFG = 1.0, CFGNorm strength = 1.0, ModelSamplingAuraFlowShift with shift = 3.0, a simple scheduler, and KSampler denoise = 1.0. The KSampler seed was randomized after each generation to increase output diversity, while workflow configuration and parameter settings were kept fixed across the dataset construction process.

\begin{table*}[!t]
\caption{Instruction-Based Generation Parameters for the Qwen Image Edit Pipeline}
\label{tab:qwen_generation_parameters}
\centering

\renewcommand{\arraystretch}{1.15}
\setlength{\tabcolsep}{5pt}

\begin{tabularx}{\textwidth}{
>{\raggedright\arraybackslash}p{0.27\textwidth}
>{\centering\arraybackslash}p{0.14\textwidth}
>{\raggedright\arraybackslash}X
}
\toprule
\textbf{Parameter} & \textbf{Value} & \textbf{Selection rationale} \\
\midrule

Output resolution
&
\(1024 \times 1024\)
&
Standardized the reference images and generated outputs for consistent quality and comparable results across scenarios.
\\

ModelSamplingAuraFlowShift
&
3.0
&
Applied as part of the adopted Qwen Image Edit ComfyUI sampling configuration.
\\

CFGNorm strength
&
1.0
&
Used to stabilize classifier-free guidance during instruction-conditioned editing.
\\

Sampling steps
&
7
&
Selected to balance visual quality and runtime efficiency.
\\

CFG scale
&
1.0
&
Selected to balance instruction adherence and visual stability.
\\

Sampler
&
Euler
&
Used as a stable and efficient diffusion sampler in the adopted workflow.
\\

Scheduler
&
Simple
&
Adopted as configured in the KSampler workflow.
\\

KSampler denoise
&
1.0
&
Used as the denoising setting in the adopted editing workflow to support the intended image edit.
\\

Seed
&
Randomized
&
Used to increase output diversity
\\

\bottomrule
\end{tabularx}
\end{table*}

Therefore, the Qwen Image Edit workflow was used as the reference-guided candidate pipeline in the comparative evaluation. Its role in the study was to test whether instruction-based editing from real Hajj reference images could provide more controlled synthetic image generation than the LoRA-adapted diffusion baselines. The final selection of this pipeline for constructing CrowdH-Edit is reported in Section 4 based on quantitative and qualitative comparison.

\subsection{Instruction Design and Controlled Editing Factors}

The proposed Qwen Image Edit pipeline relies on structured textual instructions to generate controlled synthetic variations from real Hajj reference images. Unlike open-ended text-to-image prompts, the instructions were formulated as editing commands applied to an existing reference image. Each instruction was designed to modify a specific visual factor while constraining the model to preserve the original scene structure. The main controlled factors were crowd-level condition, illumination/time-of-day condition, reference-view consistency, and location-specific preservation constraints. In this study, crowd-level conditions refer to image-level visual categories, such as very low, low, medium, and high crowd levels. This design was necessary because Hajj scenes contain sacred architectural structures and dense crowd formations, where uncontrolled generation may introduce geometry distortion, unrealistic crowd placement, or inappropriate visual artifacts.

The instruction set was developed systematically across five Hajj locations: Kaaba/Masjid al-Haram (Tawaf), Al-Jamarat, Mina, Safa-Marwah, and Arafat. The initial design space contained approximately 240 possible combinations, derived from four crowd-level conditions, four illumination conditions, three reference-view descriptions, and five locations. From this design space, 135 representative instruction templates were selected. The selection criteria included removing redundant prompt variants, prioritizing visually meaningful edits, maintaining architectural diversity across Hajj locations, and refining early trial instructions on the validation set when they produced negligible or unstable visual changes. This process provided representative coverage of the main visual conditions while reducing unnecessary generation cost.

Each instruction contains two complementary components. The positive instruction defines the intended edit, such as reducing or increasing visible crowd level, reconstructing exposed ground or floor regions, modifying illumination, or preserving the reference view. The negative instruction listed unwanted changes, including hallucinated buildings, distorted architectural elements, unrealistic body structures, blurred figures, inconsistent shadows, and color shifts in pilgrim clothing. For example, Kaaba/Tawaf instructions emphasize preservation of the Grand Mosque architecture and white marble courtyard; Al-Jamarat instructions preserve bridge geometry and safety railings; Mina instructions emphasize preservation of the tent layout; Safa-Marwah instructions preserve the walkway and rocks; and Arafat instructions preserve natural sand and rock textures.

The instruction structure was further adapted to the physical characteristics of each location. Outdoor locations used sun-position and shadow descriptions, such as early morning, midday, golden hour, or evening lighting. Indoor scenes, such as Safa-Marwah, used ceiling-light descriptions instead of sunlight-based illumination. Mina prompts were designed as lighting-only preservation cases, where the tents, crowd layout, and reference viewpoint were kept unchanged while illumination conditions were varied. Table~\ref{tab:instruction_design_dimensions} summarizes the main instruction-design dimensions used in the proposed workflow.

Table~\ref{tab:qwen_instruction_templates}  provides representative examples of the instruction templates that were used during the generation process. We selected these examples to illustrate different editing factors. The complete instruction list examples, phrase-level rationale, and location-specific prompt explanations are provided in the Supplementary Material.

\begin{table*}[!htbp]
\caption{Instruction-Design Dimensions}
\label{tab:instruction_design_dimensions}
\centering

\renewcommand{\arraystretch}{1.15}
\setlength{\tabcolsep}{5pt}

\begin{tabularx}{\textwidth}{
>{\raggedright\arraybackslash}p{0.24\textwidth}
>{\raggedright\arraybackslash}X
>{\raggedright\arraybackslash}p{0.30\textwidth}
}
\toprule
\textbf{Controlled dimension} &
\textbf{Levels / examples used} &
\textbf{Purpose} \\
\midrule

Hajj location
&
Kaaba/Masjid al-Haram (Tawaf), Al-Jamarat, Mina, Safa-Marwah, Arafat
&
Cover different pilgrimage environments and scene geometries
\\

Crowd-level condition
&
Very low, low, medium, high, and preservation only cases
&
Generate sparse, reduced, moderate, dense and unchanged crowd-layout variations
\\

Illumination / time of day
&
Early morning, midday, golden hour, evening/night
&
Introduce controlled lighting and temporal diversity
\\

Reference-view description
&
Aerial, wide-angle, ultra-wide, top-down, reference-view preservation
&
Preserve or constrain the camera perspective during editing
\\

Location-specific preservation
&
Kaaba/ Masjid al-Haram (Tawaf) architecture, Al-Jamarat railings, Mina tents, Safa-Marwah rocks, Arafat terrain
&
Preventing hallucinated or culturally inappropriate structural changes
\\

\bottomrule
\end{tabularx}
\end{table*}

%----------------------------

\begin{table*}[!htbp]
\caption{Representative Examples of Instruction Templates Used for Qwen Image Edit}
\label{tab:qwen_instruction_templates}
\centering

\renewcommand{\arraystretch}{1.15}
\setlength{\tabcolsep}{4pt}

\begin{tabularx}{\textwidth}{
>{\raggedright\arraybackslash}p{0.16\textwidth}
>{\raggedright\arraybackslash}p{0.17\textwidth}
>{\raggedright\arraybackslash}X
>{\raggedright\arraybackslash}p{0.26\textwidth}
}
\toprule
\textbf{Location} &
\textbf{Controlled factor} &
\textbf{Representative instruction summary} &
\textbf{Preservation constraint} \\
\midrule

Kaaba / Masjid al-Haram(Tawaf)
&
Very low crowd, early morning, aerial view
&
Reduce the crowd until only 10--20 pilgrims remain, reconstruct the white marble floor, and relight the scene with cool early-morning illumination and long soft shadows.
&
Do not add walls or buildings; preserve the Kaaba and Grand Mosque architecture; maintain photorealistic pilgrims and pure white clothing.
\\

Al-Jamarat
&
Medium crowd, midday, aerial view
&
Remove about half of the crowd from the bridges, reconstruct light-grey concrete floor regions, and apply bright midday lighting with short hard shadows.
&
Do not add buildings or barriers; preserve bridge geometry, safety railings, and realistic human appearance.
\\

Mina
&
Lighting-only preservation, golden hour
&
Keep the tents, crowd, layout, and viewpoint unchanged, and relight the scene with warm golden-hour illumination and long soft shadows.
&
Do not alter tent arrangement, crowd structure, road organization, or aerial viewpoint.
\\

Safa-Marwah
&
High crowd, indoor midday illumination
&
Fill the walkway with a high-density photorealistic crowd moving in two directions and simulate bright midday using neutral white ceiling lights.
&
Do not alter existing architecture; preserve the rocks and walkway structure; maintain realistic pilgrim appearance.
\\

Arafat
&
High crowd, evening, aerial view
&
Fill the plain with a high-density photorealistic crowd arranged in organic clusters and relight the scene using sparse artificial lamps.
&
Do not alter the existing landscape; preserve natural terrain and realistic pilgrim appearance.
\\

\bottomrule
\end{tabularx}
\end{table*}

\subsection{Workflow Parameter Selection}

Workflow development was performed using validation set before producing the final synthetic image sets to examine key generation parameters and identify a stable configuration for the candidate generation pipelines. The purpose of this stage was to select practical generation settings that balance instruction adherence, visual quality, and structural consistency. Table~\ref{tab:workflow_parameter_selection}  summarizes the selected parameter settings. 

The examined parameters included classifier-free guidance scale, number of inference steps, and sampler type. For the LoRA-adapted diffusion baselines, the trained LoRA adapters were loaded into the ComfyUI image-to-image workflow and guided using real Hajj reference images and textual prompts. For Qwen Image Edit, the same workflow logic was applied through reference-guided instruction-based editing. In all cases, the goal was to identify stable settings that could generate visually coherent Hajj crowd images while avoiding excessive artifacts, color distortion, or unnecessary inference cost.

The selected configuration used CFG = 1.0 and Euler sampling across the candidate pipelines. Lower CFG values weakened instruction adherence, whereas higher values increased the risk of oversaturation, color distortion, and visual artifacts. For Qwen Image Edit and SD 2.1 + LoRA, 7 inference steps provided a practical balance between output quality and runtime efficiency. For SDXL 1.0 + LoRA, 25 inference steps were selected because this setting produced more stable high-resolution outputs during validation. Euler sampling was selected because it provided stable and efficient sampling in the adopted ComfyUI workflow. These parameter-selection observations were based on qualitative inspection of the validation outputs.

\begin{table*}[!t]
\caption{Workflow Parameter Selection for Candidate Generation Pipelines}
\label{tab:workflow_parameter_selection}
\centering

\footnotesize
\renewcommand{\arraystretch}{1.15}
\setlength{\tabcolsep}{4pt}

\begin{tabularx}{\textwidth}{
>{\raggedright\arraybackslash}p{0.17\textwidth}
>{\raggedright\arraybackslash}p{0.13\textwidth}
>{\raggedright\arraybackslash}X
>{\raggedright\arraybackslash}p{0.25\textwidth}
}
\toprule
\textbf{Parameter} &
\textbf{Values Tested} &
\textbf{Observations} &
\textbf{Final Selection \& Justification} \\
\midrule

CFG Scale (Classifier Free Guidance)
&
0.5, 1.0, 1.5, 2.0
&
CFG=0.5: Weak instruction adherence\newline
CFG=1.0: Strong instruction alignment with stable visual quality\newline
CFG=1.5: Acceptable output but slight oversaturation\newline
CFG\(\geq\)2.0: Increased artifacts, color distortion, oversaturation
&
CFG=1.0 was selected for all candidate pipelines to balance instruction adherence and visual stability.
\\

Inference Steps (Diffusion iterations)
&
3, 5, 7, 10, 15, 20,25
&
Steps \(<\) 4: produced incomplete or blurry outputs\newline
Steps=7: provided stable visual quality\newline
Steps\(>\)10: Diminishing returns for Qwen Image Edit and SD 2.1 + LoRA\newline
Steps=25: Produced more stable high-resolution outputs for SDXL 1.0 + LoRA.
&
7 steps (Qwen Image Edit, SD 2.1 + LoRA)\newline
25 steps (SDXL 1.0 + LoRA)\newline
(selected as a practical quality)
\\

Sampler Type
&
Euler, Euler A, DPM++, Heun
&
Euler: Stable and visually consistent outputs.\newline
Euler A: Greater stochastic variation.\newline
DPM++: Visually coherent outputs, but no clear qualitative
advantage over Euler.\newline
Heun: Visually good outputs, but no consistent qualitative
advantage over Euler.
&
Euler was selected for all candidate pipelines because it provided stable and visually consistent outputs under the adopted workflow.
\\

\bottomrule
\end{tabularx}
\end{table*}

\subsection{Synthetic Dataset Generation}
After the model and workflow settings were fixed, the 261 held-out real Hajj reference images, organized by location, were used as the source pool for candidate generation. The 135 instruction templates comprised 27 location-specific templates for each of the five Hajj locations. During batch generation, the corresponding location-specific instructions were applied across the reference images, while the generation seed was randomized and the remaining pipeline configuration was kept fixed. Generation runs were repeated across locations and instruction conditions until a candidate pool of 8,000 images was obtained.

Generated images were organized using a structured filename convention containing the location label, generation method, and serial image identifier. This convention supported grouping by location and pipeline during evaluation and filtering. All outputs were initially treated as candidate images requiring quality screening before inclusion in the final dataset. Potential issues included visual artifacts, duplicate content, geometric inconsistencies, and unsuitable crowd regions.

\subsection{Quantitative Image Evaluation}

To compare the three generation pipelines under matched conditions, a separate controlled evaluation set was generated independently of the 8,000-image candidate pools. The same 261 hold out test images and corresponding editing conditions were used for all three pipelines. For each reference--instruction case, one output was generated by each pipeline using its validation-selected inference configuration. No manual best-sample selection or post-generation quality filtering was applied before metric computation. This yielded 261 matched reference--output pairs per pipeline. The evaluation assessed reference-image preservation, distributional similarity, and image–text alignment.

Distributional fidelity was assessed using the Fréchet Inception Distance (FID) \cite{b63,b64} and Kernel Inception Distance (KID) \cite{b65}. The same 261 real images formed the comparison set for each pipeline’s 261 generated outputs. Features were extracted using ImageNet-pretrained Keras InceptionV3 with global average pooling, producing 2,048-dimensional representations. Input images were represented as float32 values in [0, 255], resized to 299 × 299 pixels using bilinear interpolation with antialiasing, and processed using the InceptionV3 preprocessing function. FID was calculated from the means and covariance matrices of the real and generated feature distributions.
KID was computed using a NumPy implementation of the unbiased
squared maximum mean discrepancy estimator with the polynomial
kernel \(k(x,y) = (x^{\top}y/d + 1)^3\), where \(d = 2{,}048\). For each of 50 repetitions, subsets of 200 images were independently sampled from the real and generated collections, without replacement within each subset. KID was reported as the mean and standard deviation across the 50 estimates. The standard deviation was calculated using ddof = 0, and the subset-sampling seed was fixed at 1337. This standard deviation describes variation across the sampled subsets. Lower FID and KID values indicate closer distributional agreement under the specified feature-extraction protocol.

Reference-image preservation was assessed using the Structural Similarity Index (SSIM) \cite{b67}, Peak Signal-to-Noise Ratio (PSNR) \cite{b66}, and Learned Perceptual Image Patch Similarity (LPIPS) \cite{b68}. Each generated image was compared with the real image used as its generation reference. For these paired comparisons, images were converted to RGB and resized to 768 × 768 pixels using bilinear interpolation with antialiasing. SSIM and PSNR were computed using TensorFlow with pixel values scaled to [0, 1]. LPIPS was computed using the VGG backbone with inputs normalized to $[-1,1]$. Each metric was averaged across the 261 pairs, and PSNR was reported in decibels. Higher SSIM and PSNR and lower LPIPS indicate closer agreement with the reference. Because the task involves controlled editing, these measures were interpreted as indicators of reference preservation rather than direct measures of editing success. Semantic alignment was assessed using the TorchMetrics implementation of CLIPScore with pretrained CLIP ViT-B/32 \cite{b69}. Each generated image was compared with a corresponding target description derived from the editing instructions. Scores were reported on the 0--100 scale as mean $\pm$ standard deviation across 261 image--text pairs per pipeline.

\subsection{Pipeline Comparison and Final Selection}

Using the controlled evaluation protocol described in Subsection 3.7, we compared SD 2.1 + LoRA, SDXL 1.0 + LoRA, and Qwen Image Edit to select the pipeline for constructing CrowdH-Edit. This stage was designed to assess not only visual realism but also the suitability of each pipeline for producing Hajj crowd images that could support crowd-counting research.

The quantitative results were complemented by a qualitative visual comparison of outputs from the three candidate pipelines. This review focused on the preservation of sacred-site geometry, realistic pilgrim appearance, consistent illumination, absence of hallucinated structures, and appropriate visual appearance. These checks addressed Hajj-specific visual requirements that automatic metrics may not fully capture.
The final selection balanced quantitative performance, visual realism, structural consistency, and coverage of the intended visual conditions. The candidate image pool produced by the selected pipeline was subsequently cleaned to construct the CrowdH-Edit dataset. The remaining pipelines were retained as comparative baselines.  Detailed numerical comparisons and qualitative examples are presented in Section 4.1.

\subsection{Data Cleaning and Human Qualitative Screening}

After selecting Qwen Image Edit as the final generation pipeline, we cleaned its candidate pool of 8,000 images. CleanVision-based inspection \cite{b70} was used to identify technical image-quality issues, including low-information images, blurred outputs, duplicate or near-duplicate samples, and visually degraded images. Images with poor technical quality or redundant content were removed during this stage.
The remaining images were manually reviewed to assess their suitability for crowd-counting annotation and model training. The review examined crowd clarity, the visibility of individual people and head regions, and the consistency of the generated scenes with the Hajj context. Images were excluded if they contained severe crowd blur, merged or distorted bodies, unclear crowd patterns, unrealistic pedestrian placement, geometry errors, hallucinated structures, or crowd regions unsuitable for reliable head-point annotation. Following this two-stage screening process, 6,122 images were retained as the cleaned CrowdH-Edit dataset. This unannotated image collection was used for dataset-level analysis and the subsequent selection of samples for annotation.

\subsection{CrowdH-Edit-Ann-387 Annotation Protocol }

The cleaned CrowdH-Edit dataset was used as the synthetic image pool from which the annotated subset was selected. A subset of 387 images was selected based on visual quality, clarity of head regions, and coverage of different Hajj locations, crowd-level conditions, illumination settings, and scene layouts. This subset was annotated using head-point annotations in the VGG Image Annotator (VIA) tool \cite{b71} to construct CrowdH-Edit-Ann-387. The annotation guideline was to mark the center of each clearly visible pilgrim head. We labeled only visible and distinguishable heads to avoid estimating hidden, severely blurred, merged, or ambiguous people. The quality of the annotation was verified through repeated manual review. Any missed, duplicated, or incorrectly placed points identified during verification were corrected before finalizing the dataset.

The separation between the cleaned synthetic pool and the annotated subset enables two levels of use. The full CrowdH-Edit dataset supports dataset-level visual analysis and future sample selection, whereas CrowdH-Edit-Ann-387 provides head-point annotations for supervised downstream crowd-counting validation. This strategy supports our objective of producing a Hajj-specific synthetic dataset that expands visual diversity while maintaining annotation reliability for crowd-counting research.

\subsection{Properties of CrowdH-Edit-Ann-387}

CrowdH-Edit-Ann-387 contains 387 synthetic Hajj crowd images with a resolution of 1037 × 1063. The dataset contains 49,375 annotated pedestrians. Table~\ref{tab:dataset_statistical_comparison} compares CrowdH-Edit-Ann-387 with representative real and synthetic crowd-counting datasets. Although large synthetic datasets such as GCC and CrowdX provide substantially greater scale through simulation, they are generated in virtual environments and do not explicitly model Hajj-specific scene geometry or pilgrimage appearance. In contrast, CrowdH-Edit-Ann-387 is smaller, but it is designed for a different purpose: providing an annotated, Hajj-specific synthetic subset generated through reference-guided image editing. Similarly, real benchmarks such as ShanghaiTech, UCF-QNRF, and NWPU-Crowd provide important general-purpose resources, but they do not explicitly represent Hajj-specific conditions. Therefore, the value of CrowdH-Edit-Ann-387 lies in its target-domain relevance and its role as a synthetic training-support subset for Hajj crowd-counting research.

\begin{table*}[!t]
\caption{Statistical Comparison between Representative Crowd-Counting Datasets and the Proposed CrowdH-Edit-Ann-387 Dataset}
\label{tab:dataset_statistical_comparison}
\centering
\footnotesize
\renewcommand{\arraystretch}{1.15}
\setlength{\tabcolsep}{3pt}

\begin{tabularx}{\textwidth}{
>{\raggedright\arraybackslash}p{0.19\textwidth}
>{\centering\arraybackslash}p{0.14\textwidth}
>{\centering\arraybackslash}p{0.085\textwidth}
>{\centering\arraybackslash}p{0.115\textwidth}
>{\centering\arraybackslash}p{0.065\textwidth}
>{\centering\arraybackslash}p{0.065\textwidth}
>{\centering\arraybackslash}p{0.075\textwidth}
>{\centering\arraybackslash}X
}
\toprule
\textbf{Dataset} &
\textbf{Avg. img. size} &
\textbf{\# Images} &
\textbf{Total count} &
\textbf{Min} &
\textbf{Avg} &
\textbf{Max} &
\textbf{Type} \\
\midrule

GCC~\cite{b16}
& \(1080 \times 1920\) & 15,212 & 7,625,843 & 0 & 501 & 3,995 & Synth \\

CVCS~\cite{b44}
& \(1080 \times 1920\) & 280,000 & -- & 90 & -- & 180 & Synth \\

CrowdX~\cite{b17}
& \(1024 \times 768\) & 24,000 & -- & 1 & 500 & 1,000 & Synth \\

SHT A~\cite{b6}
& \(589 \times 868\) & 482 & 241,677 & 33 & 501 & 3,139 & Real \\

SHT B~\cite{b6}
& \(768 \times 1024\) & 716 & 88,488 & 9 & 123 & 578 & Real \\

WorldExpo'10~\cite{b13}
& \(576 \times 720\) & 3,980 & 199,923 & 1 & 50 & 253 & Real \\

UCF-QNRF~\cite{b41}
& \(2013 \times 2902\) & 1,535 & 1,251,642 & 49 & 815 & 12,865 & Real \\

NWPU~\cite{b43}
& \(2191 \times 3209\) & 5,109 & 2,133,375 & 0 & 418 & 20,033 & Real \\

CrowdH-Edit-Ann-387 (Our)
& \(1037 \times 1063\) & 387 & 49,375 & 2 & 128 & 2,923 & Synth \\

\bottomrule
\end{tabularx}
\end{table*}

\begin{figure}[!htbp]
\centering
\includegraphics[width=\columnwidth]{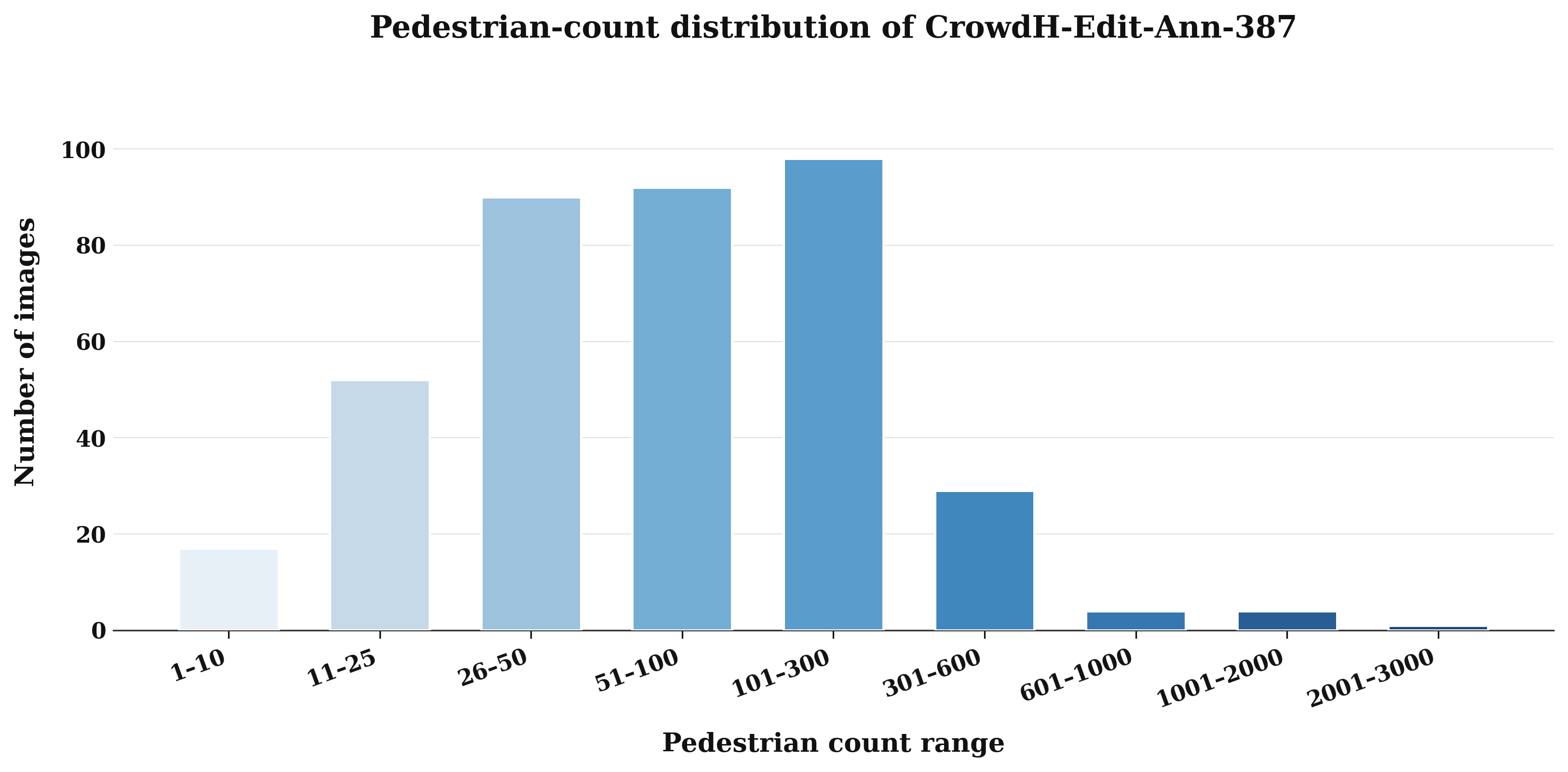}
\caption{Scene-category distribution of CrowdH-Edit-Ann-387.}
\label{fig:Fig_3}
\end{figure}

\begin{figure}[!htbp]
\centering
\includegraphics[width=\columnwidth]{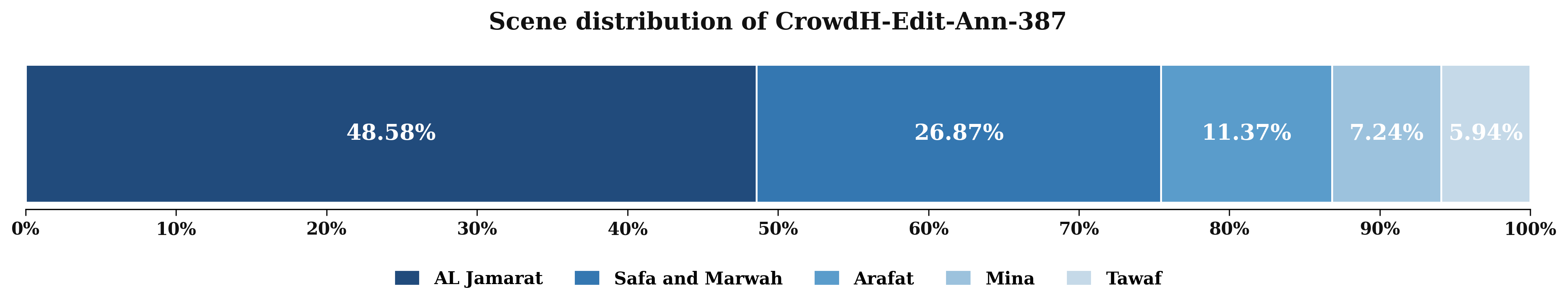}
\caption{Pedestrian-count distribution of CrowdH-Edit-Ann-387.}
\label{fig:Fig_4}
\end{figure}

Figure ~\ref{fig:Fig_3} shows the image-level pedestrian-count distribution of CrowdH-Edit-Ann-387. The dataset covers sparse, low-count, moderate, and high-count Hajj crowd scenes, with counts ranging from 2 to 2,923 pedestrians per image. Figure ~\ref{fig:Fig_4} shows the scene-category distribution across the main Hajj locations represented in the annotated subset, including Al-Jamarat, Safa-Marwah, Arafat, Tawaf/Kaaba, and Mina. These distributions indicate that the annotated subset provides both count-level and location-level variation. This coverage is important because Hajj crowd-counting models must generalize across different spatial layouts as bridge structures, sacred courtyards, indoor walkways, tent-road environments, and open pilgrimage scenes. Beyond count-level and location-level coverage, CrowdH-Edit-Ann-387 was designed to provide controlled visual diversity through reference-guided instruction-based editing. The generation instructions targeted variations in Hajj location, crowd-level condition, illumination/time-of-day setting, and reference-view preservation. Qualitative examples of these controlled variations are presented in subsection 4.2, where Figure ~\ref{fig:Fig_8} illustrates location-level diversity, Figure ~\ref{fig:Fig_9} demonstrates crowd-level control, and Figure ~\ref{fig:Fig_10} shows illumination/time-of-day variation.

\subsection{Crowd-Counting Validation }

The usefulness of CrowdH-Edit-Ann-387 for crowd counting was evaluated using MCNN \cite{b6}, CSRNet \cite{b72}, DM-Count \cite{b73}, P2PNet \cite{b74}, and APGCC \cite{b75}. These models cover density-regression, distribution-aware, and point/localization-based approaches, allowing the effect of synthetic augmentation to be examined across different counting models.

The models were implemented using their model-specific configurations, while the original architecture and optimization settings were retained. For this study, the data protocol was standardized using the same five-fold partitions and the same fixed real test set across all models. Within each model, the training and preprocessing configurations were kept unchanged between the real-only and real-plus-synthetic settings; the only difference was the inclusion of CrowdH-Edit-Ann-387 in the training folds.

The evaluation used 307 real Hajj images for development and a separate set of 77 real images for testing \cite{b57}. In the augmented setting, the 387 annotated synthetic images were added to the training data, giving 694 development images in total. The 307 real images were divided into five folds stratified by annotated person count, with 245 or 246 real images used for training and 61 or 62 for validation in each fold. Under synthetic augmentation, all 387 synthetic images were added only to the training subset, producing 632 or 633 training images per fold, while validation remained restricted to real images. No synthetic images were included in the fixed 77-image test set, and none of the synthetic training images were generated from these test images. Table ~\ref{tab:downstream_validation_subsets} summarizes these data subsets. 

For each fold, the checkpoint with the lowest validation Mean Absolute Error (MAE) was selected and evaluated on the fixed 77-image test set. Performance was measured using MAE and Root Mean Squared Error (RMSE) \cite{b72} and the results were reported as mean $\pm$ sample standard deviation across the five folds.

\begin{table*}[!t]
\caption{Data Subsets Used in Downstream Crowd-Counting Validation}
\label{tab:downstream_validation_subsets}
\centering
\footnotesize
\renewcommand{\arraystretch}{1.15}
\setlength{\tabcolsep}{3pt}

\begin{tabularx}{\textwidth}{
>{\raggedright\arraybackslash}p{0.19\textwidth}
>{\centering\arraybackslash}p{0.13\textwidth}
>{\centering\arraybackslash}p{0.075\textwidth}
>{\centering\arraybackslash}p{0.10\textwidth}
>{\centering\arraybackslash}p{0.055\textwidth}
>{\centering\arraybackslash}p{0.055\textwidth}
>{\centering\arraybackslash}p{0.075\textwidth}
>{\centering\arraybackslash}p{0.065\textwidth}
>{\centering\arraybackslash}X
}
\toprule
\textbf{Dataset}
& \textbf{Avg. img. size}
& \textbf{\# Images}
& \textbf{Total count}
& \textbf{Min}
& \textbf{Avg}
& \textbf{Max}
& \textbf{Type}
& \textbf{Target Domain} \\
\midrule

Real-Hajj Data \cite{b57}
& \(1076 \times 1647\)
& 307
& 180349
& 12
& 587
& 5665
& Real
& Hajj \\

CrowdH-Edit-Ann-387(our)
& \(1037 \times 1063\)
& 387
& 49375
& 2
& 128
& 2923
& Synth
& Hajj \\

Real-Hajj Data + CrowdH-Edit-Ann-387
& \(1054 \times 1322\)
& 694
& 229724
& 2
& 331
& 5665
& Mixed
& Hajj \\

Real-Test \cite{b57}
& \(1099 \times 1590\)
& 77
& 31207
& 70
& 405
& 5097
& Real
& Hajj \\

\bottomrule
\end{tabularx}
\end{table*}

\begin{figure}[!htbp]
\centering
\includegraphics[width=\columnwidth]{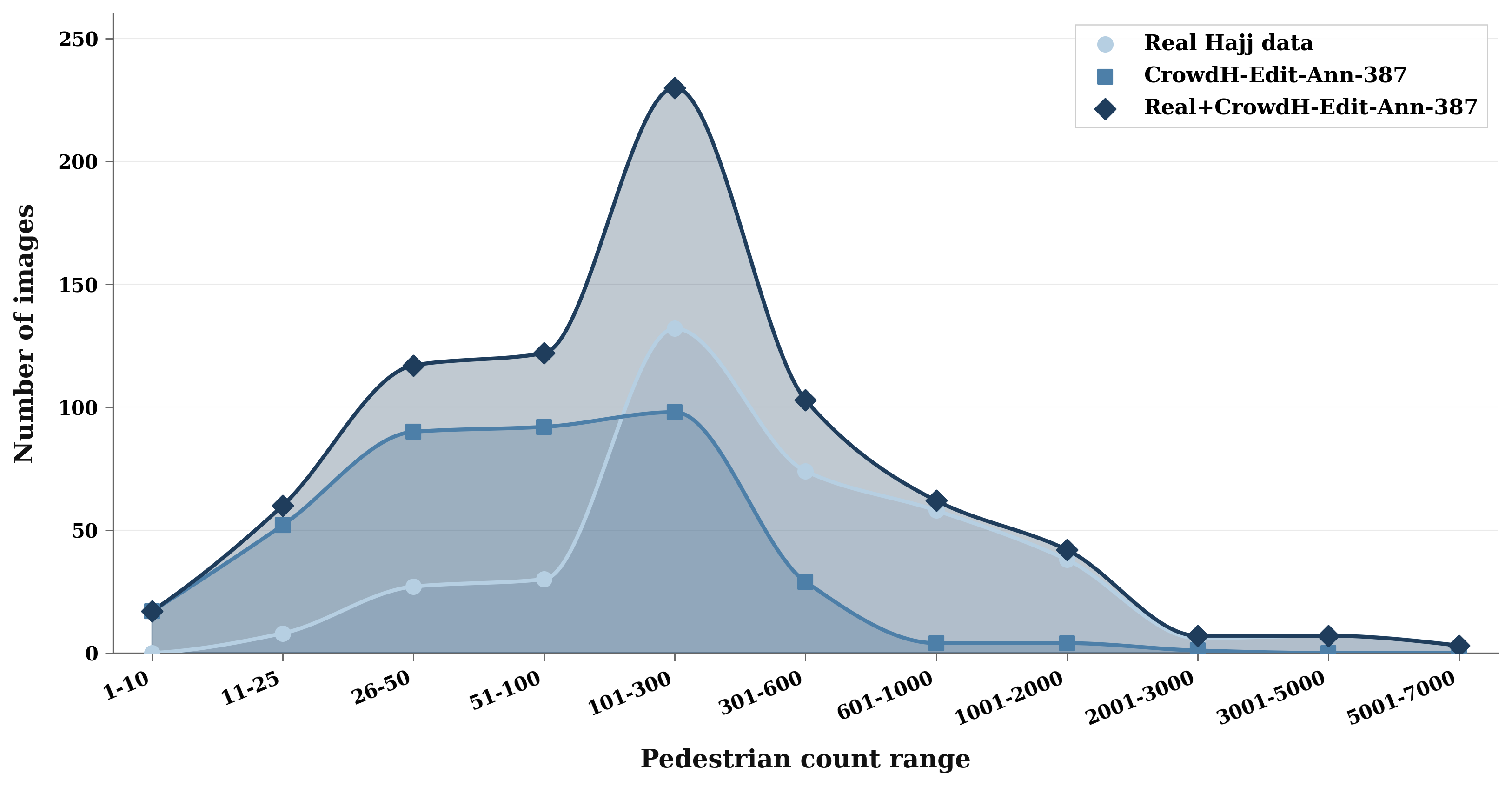}
\caption{Pedestrian-count distribution of the Real Hajj data, CrowdH-Edit-Ann-387, and Real+CrowdH-Edit-Ann-387.}
\label{fig:Fig_5}
\end{figure}

The pedestrian-count distributions of the Real-Hajj dataset, CrowdH-Edit-Ann-387, and the combined real-plus-synthetic training set are shown in Figure~\ref{fig:Fig_5}. The Real-Hajj Data contain a larger proportion of medium-to-high count scenes, whereas CrowdH-Edit-Ann-387 contributes additional samples across sparse and low-to-high count conditions. This complementary distribution broadens the range of image-level crowd counts represented in the training data and supports the real-only versus real-plus-synthetic validation protocol.

\begin{figure}[!htbp]
\centering
\includegraphics[width=\columnwidth]{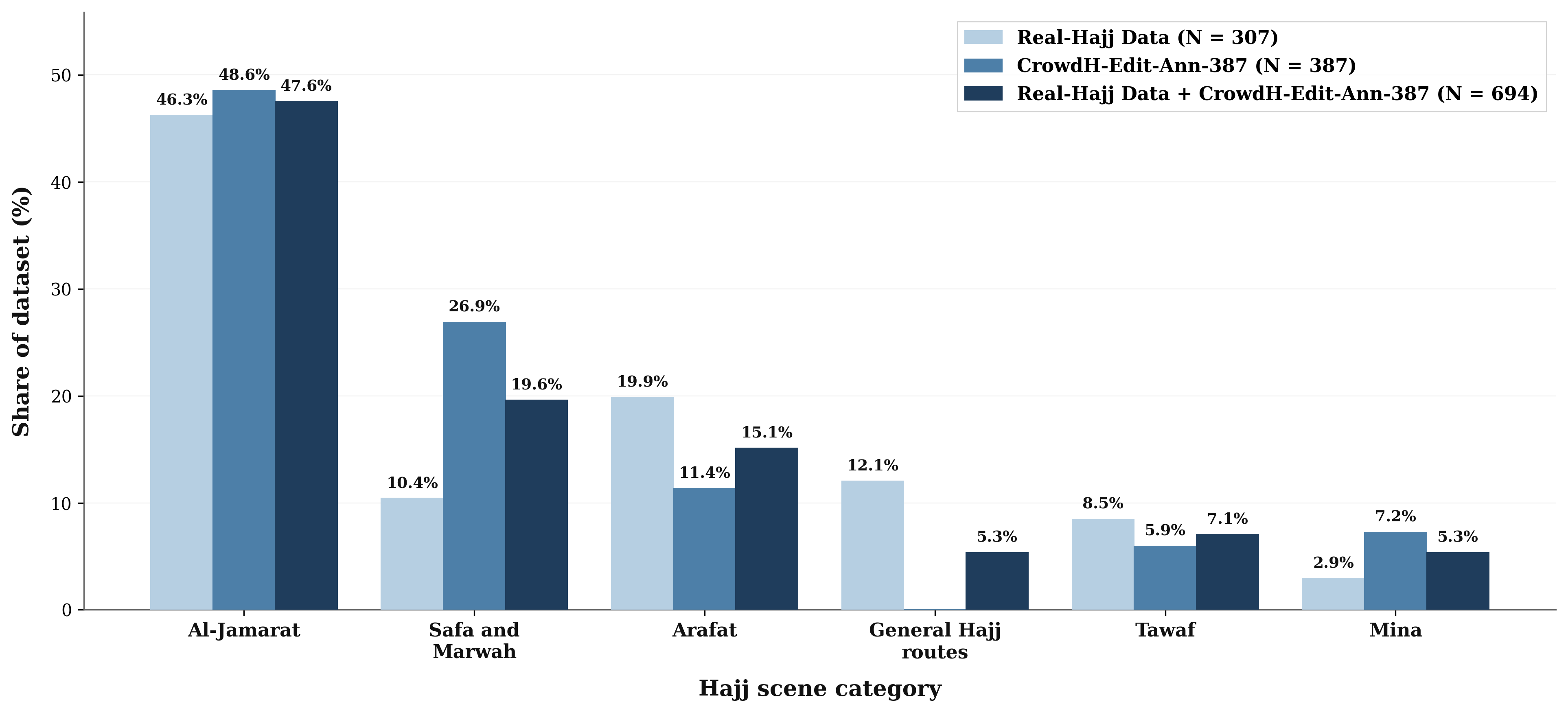}
\caption{Scene-category distribution of the Real Hajj data,
CrowdH-Edit-Ann-387, and the combined real-plus-synthetic training set.}
\label{fig:Fig_6}
\end{figure}

The scene-category distribution differs between the Real-Hajj dataset, CrowdH-Edit-Ann-387, and the combined real-plus-synthetic training set, as shown in Figure~\ref{fig:Fig_6}. The added synthetic subset increases the representation of Safa-Marwah and Mina while maintaining strong coverage of Al-Jamarat. This distribution supports the validation protocol by expanding the scene composition of the training data across multiple Hajj locations.

\section{ Results and Discussion}

\subsection{Quantitative and Qualitative Comparison of Candidate Generation Pipelines}

Qwen Image Edit achieved the best scores across the six reported metrics as illustrated in Table \ref{tab:candidate_pipeline_comparison}. Its outputs showed stronger image–text alignment, closer agreement with the real-image distribution, and greater similarity to the reference images than the two LoRA-based baselines. The agreement across these complementary measures supports its suitability for reference-guided Hajj image generation under the evaluated conditions.
The comparison between the diffusion baselines reveals a less uniform improvement. SDXL 1.0 + LoRA achieved lower FID and KID and higher CLIPScore, SSIM, and PSNR than SD 2.1 + LoRA. However, LPIPS increased from 0.7593 to 0.7902, indicating weaker perceptual similarity to the references. These results show that improved distributional similarity does not necessarily imply stronger preservation of reference-image content. For geometry-sensitive Hajj imagery, this distinction is important because the generated images should remain realistic while preserving scene elements that are not targeted by the editing instructions.

\begin{table*}[!t]
\caption{Quantitative Comparison of Candidate Generation Pipelines}
\label{tab:candidate_pipeline_comparison}
\centering
\footnotesize
\renewcommand{\arraystretch}{1.15}
\setlength{\tabcolsep}{3pt}

\begin{tabularx}{\textwidth}{
>{\raggedright\arraybackslash}p{0.18\textwidth}
>{\centering\arraybackslash}p{0.16\textwidth}
>{\centering\arraybackslash}p{0.08\textwidth}
>{\centering\arraybackslash}p{0.08\textwidth}
>{\centering\arraybackslash}p{0.12\textwidth}
>{\centering\arraybackslash}p{0.09\textwidth}
>{\centering\arraybackslash}X
}
\toprule
\textbf{Pipeline} &
\textbf{CLIPScore $\pm$ Std $\uparrow$} &
\textbf{FID $\downarrow$} &
\textbf{SSIM $\uparrow$} &
\textbf{PSNR (dB) $\uparrow$} &
\textbf{LPIPS $\downarrow$} &
\textbf{KID mean $\pm$ Std $\downarrow$} \\
\midrule

SD 2.1 + LoRA &
$61.87 \pm 5.18$ &
287.51 &
0.1245 &
9.2580 &
0.7593 &
$0.321 \pm 0.0106$ \\

SDXL 1.0 + LoRA &
$67.24 \pm 9.47$ &
117.81 &
0.2109 &
9.4140 &
0.7902 &
$0.103 \pm 0.014$ \\

Qwen Image Edit &
$92.25 \pm 5.37$ &
93.85 &
0.7345 &
18.61 &
0.2658 &
$0.0334 \pm 0.0059$ \\

\bottomrule
\end{tabularx}
\end{table*}

\begin{figure}[!htbp]
\centering
\includegraphics[width=\columnwidth]{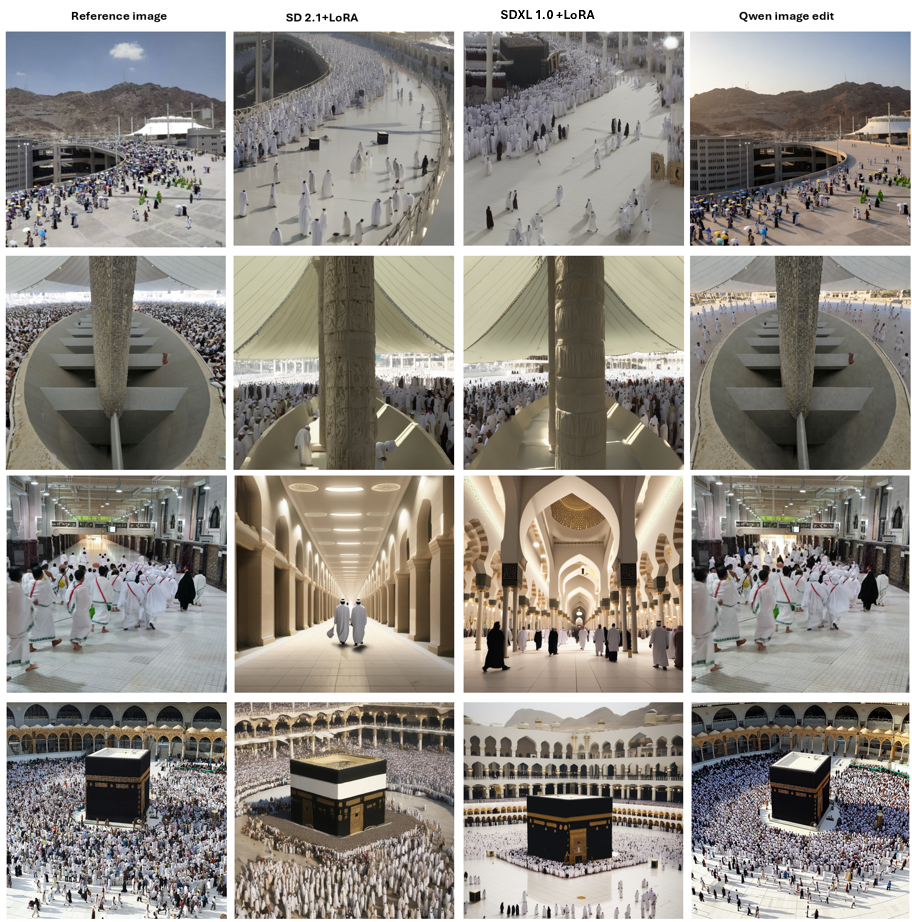}
\caption{Visual comparison of candidate generation pipelines. Each row shows a real Hajj reference image and the corresponding outputs generated by SD 2.1 + LoRA, SDXL 1.0 + LoRA, and Qwen Image Edit.}
\label{fig:Fig_7}
\end{figure}

The visual examples in Figure \ref{fig:Fig_7} provide further evidence of this difference. The LoRA-based outputs introduce substantial changes in viewpoint and architectural arrangement, particularly in the Al-Jamarat and indoor walkway examples. Qwen Image Edit retains a closer correspondence with the reference scenes, including the arrangement of bridge structures, walkway features, and the Kaaba surroundings. These observations complement the quantitative results by identifying scene changes that are relevant to the preservation of Hajj-specific visual context. Therefore, the quantitative results and visual comparisons supported the selection of Qwen Image Edit to construct CrowdH-Edit.

%,height=0.36\textheight

\subsection{Instruction-Controlled Qualitative Analysis}
 
\begin{figure}[!htbp]
\centering
\includegraphics[width=\columnwidth]{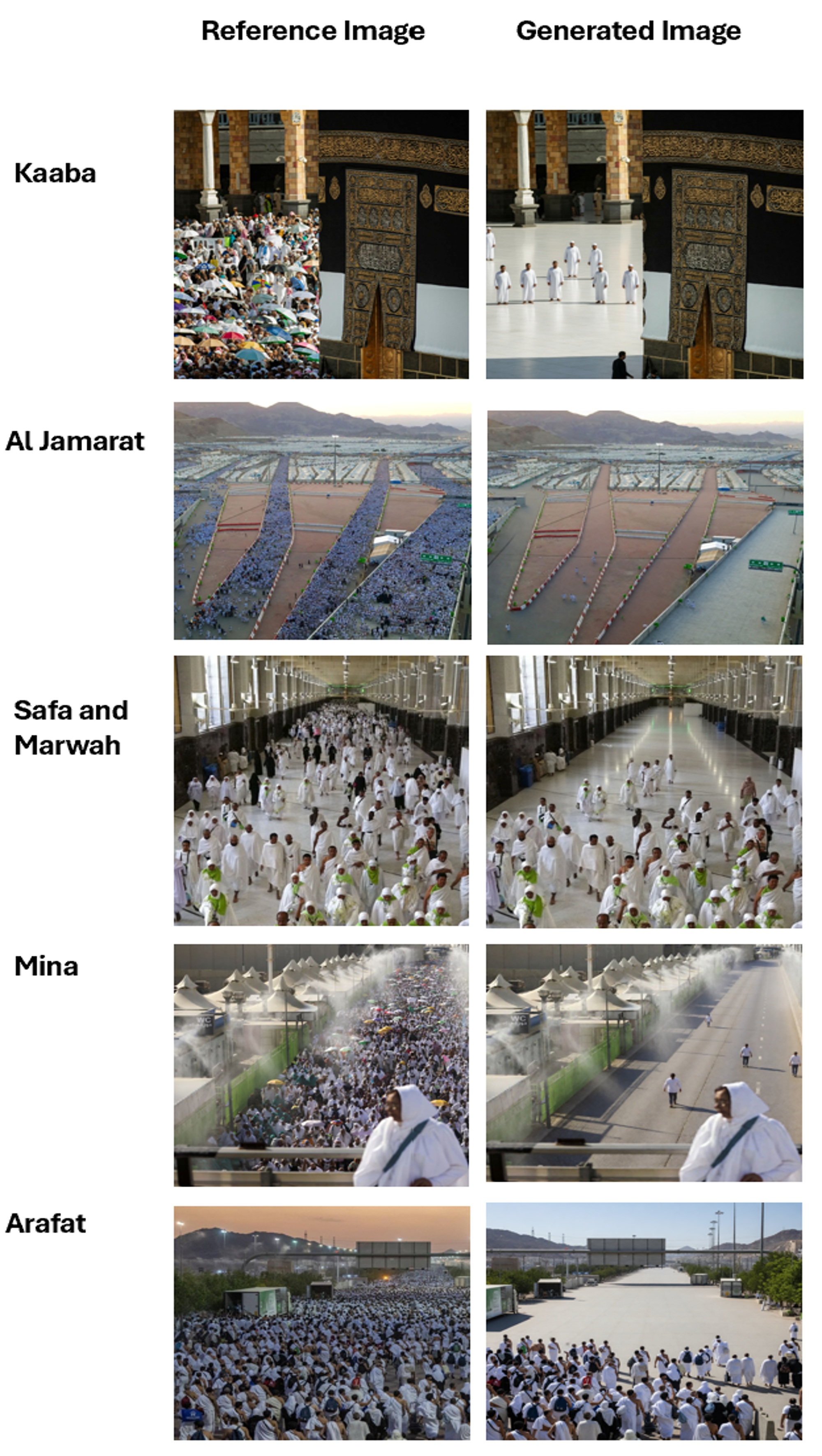}
\caption{Cross-location qualitative comparison between real Hajj reference images and Qwen Image Edit outputs. Each row shows a representative real reference image and its corresponding generated output for one Hajj location. The examples illustrate that the proposed workflow can modify visible crowd conditions while  preserving location-specific structures, including Kaaba/Tawaf surroundings, Al-Jamarat bridge layout, Safa-Marwah walkways, Mina tent-road environments, and Arafat terrain context.}
\label{fig:Fig_8}
\end{figure}

\begin{figure}[!htbp]
\centering
\includegraphics[width=\columnwidth]{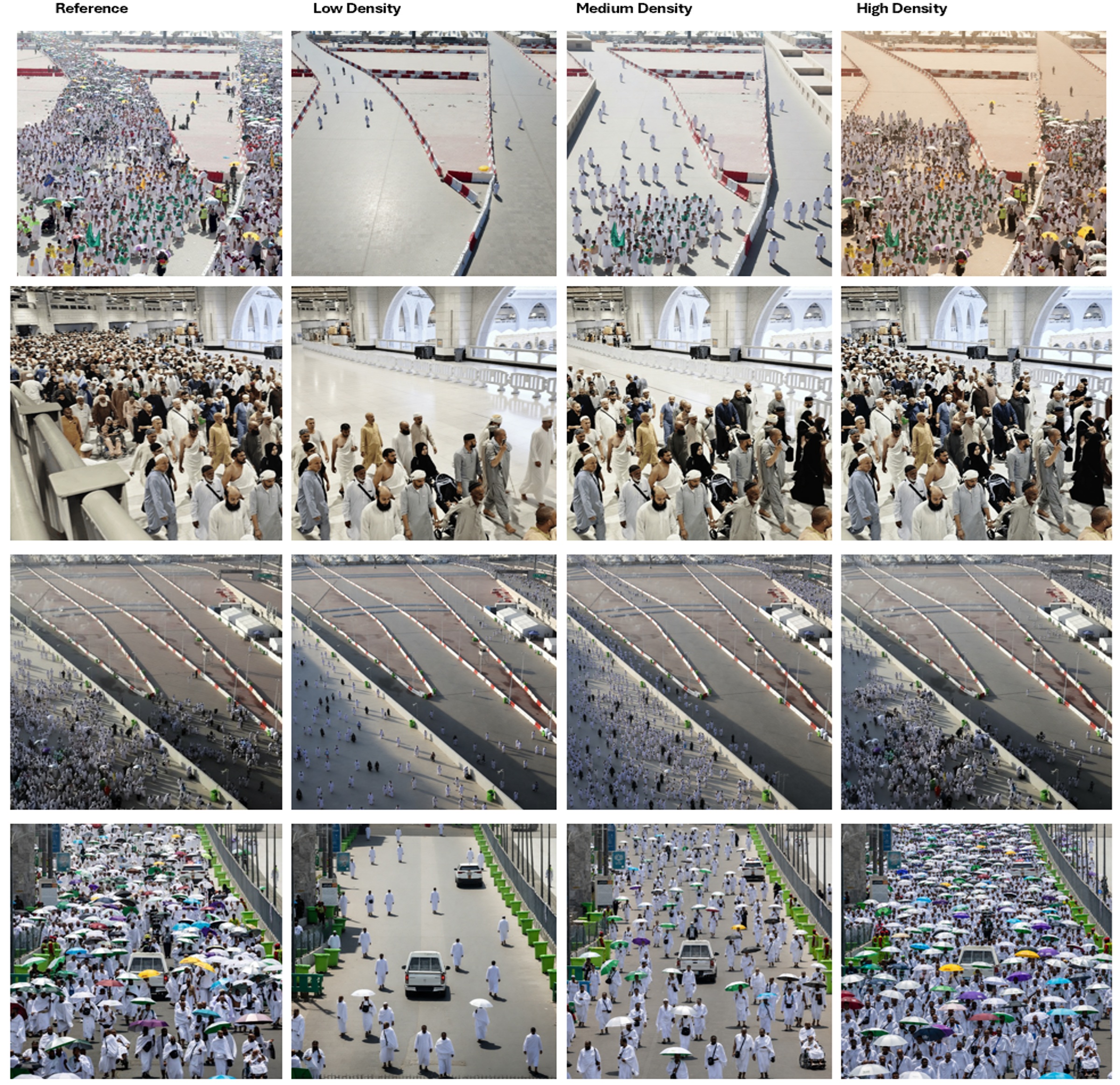}
\caption{Crowd-level control examples generated using Qwen Image Edit. Each row shows a real Hajj reference image followed by generated outputs under low, medium, and high crowd-level conditions. The examples show that the workflow can modify the number and spatial distribution of visible pilgrims while largely preserving the main road layout, barriers, architectural context, and scene geometry.}
\label{fig:Fig_9}
\end{figure}

\begin{figure}[!htbp]
\centering
\includegraphics[width=\columnwidth]{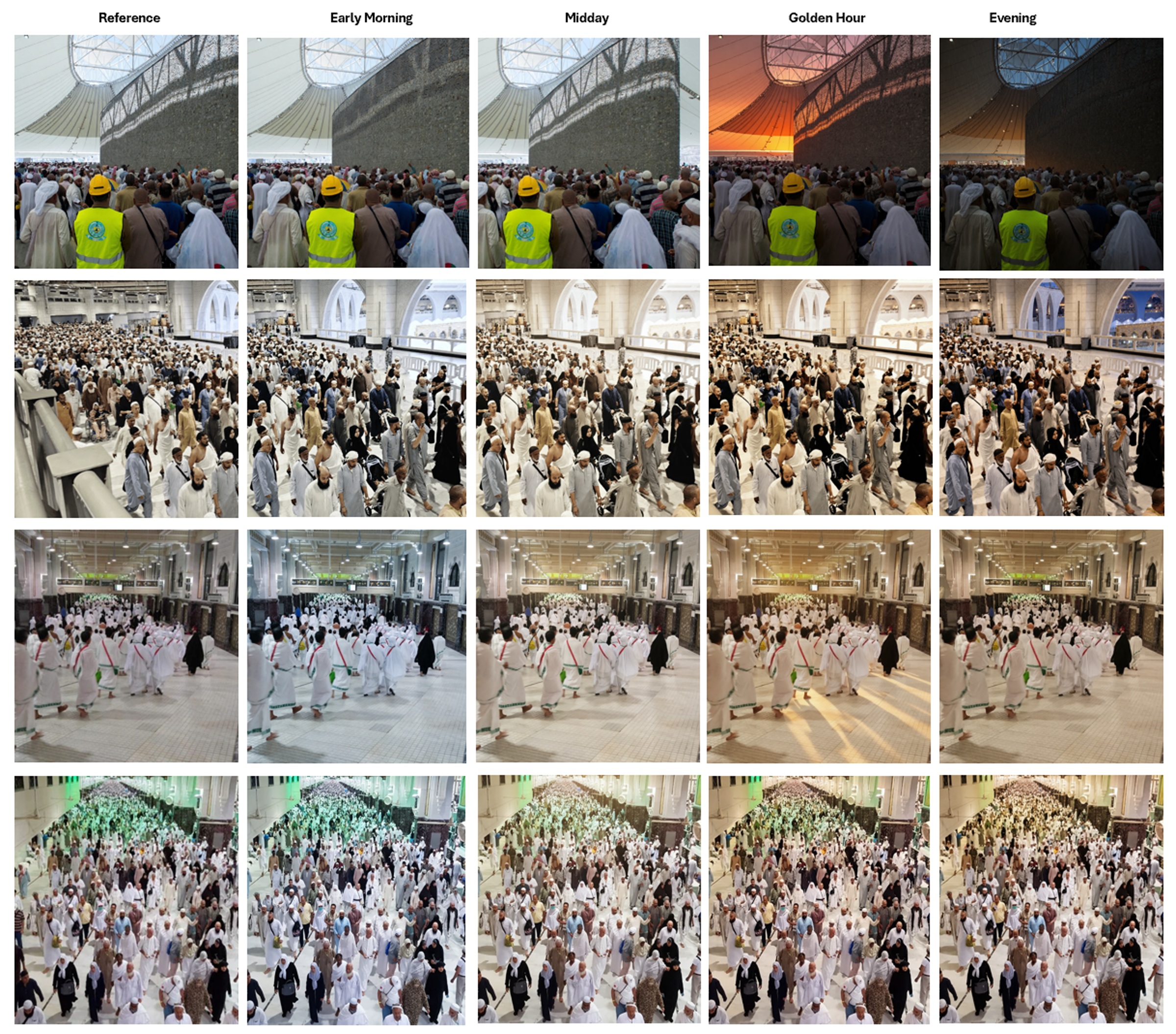}
\caption{Time-of-day and illumination control examples generated using Qwen Image Edit. Each row shows a real Hajj reference image followed by edited outputs under early-morning, midday, golden-hour, and evening illumination conditions. The examples demonstrate that the proposed workflow can modify brightness, color temperature, shadow behavior, and scene mood while largely preserving the main scene structure and overall viewpoint.}
\label{fig:Fig_10}
\end{figure}

 The qualitative assessment was conducted to examine whether the generated images satisfy Hajj-specific visual needs that are not fully captured through quantitative metrics. The analysis focused on three aspects: cross-location consistency, crowd-level control, and illumination/time-of-day variation. Representative examples are presented in Figures ~\ref{fig:Fig_8}, ~\ref{fig:Fig_9}, and ~\ref{fig:Fig_10} to demonstrate the main controllability properties of the proposed Qwen Image Edit workflow.

Figure ~\ref{fig:Fig_8} presents cross-location examples that compare real Hajj reference images with their corresponding Qwen Image Edit outputs. The generated images retain the main visual identity of each location while applying the specified edits. In the Kaaba/Tawaf examples, the sacred structure and surrounding architectural elements remain recognizable after crowd modification. In Al-Jamarat, the bridge lanes, barriers, and surrounding infrastructure are largely preserved. In Safa-Marwah, the indoor walkway and architectural context have been consistently maintained. In Mina and Arafat, the tent-road environment, open terrain, and large-scale spatial organization remain visually recognizable. These examples illustrate that a reference-guided editing strategy can produce location-aware synthetic Hajj scenes while maintaining key structural cues.

The examples of crowd control at the low, medium and high crowd conditions are illustrated in Figure~\ref{fig:Fig_9}. The low-crowd output exposes larger areas of the floor, road, or ground. The medium- and high-crowd outputs show progressively greater crowd presence relative to the low-crowd outputs. These levels describe the intended editing conditions, so a high-crowd output may still be less crowded than its original reference image. This variation supports dataset construction by providing images with different visible crowd levels. However, their suitability for reliable head-point annotation still depends on clearly distinguishable heads.

Figure~\ref{fig:Fig_10} shows the effect of illumination and time-of-day instructions. Across the illustrated examples, the main viewpoint, architectural layout, and overall crowd arrangement remain largely recognizable. The lighting changes are more pronounced in some scenes than in others. These observations support the workflow’s ability to vary illumination while largely retaining the reference scene layout. At the same time, the generated images retain the main viewpoint, architectural structure, and crowd organization. This indicates that the workflow can introduce temporal and lighting variation without altering the underlying Hajj scene structure.

Overall, the qualitative results show that Qwen Image Edit offers controllable visual variation across Hajj locations, crowd-level conditions, and illumination settings. This is important for downstream crowd-counting research because generated images must not only appear realistic, but also preserve scene geometry, maintain plausible crowd placement, and provide sufficiently clear people and head regions for reliable annotation and model training.

\subsection{Crowd-Counting Validation}
 
Adding CrowdH-Edit-Ann-387 improved counting performance for some models as shown in Table~\ref{tab:downstream_counting_performance}. However, the magnitude and direction of the changes varied across models and metrics. MCNN showed the largest reduction in MAE, accompanied by only modest decrease in RMSE. The substantially larger reduction in MAE than in RMSE suggests that augmentation reduced the typical absolute counting errors more effectively than the relatively large errors emphasized by RMSE. APGCC also improved in both metrics and achieved the lowest errors among the models trained with augmentation. However, its small MAE reduction coincided with greater variability in the reported MAE scores. These differences highlight the importance of considering both the magnitude and variability of the observed gains.

DM-Count showed a mixed response: its MAE decreased, whereas its RMSE increased. The lower MAE therefore did not indicate improvement across both error measures. The higher RMSE also points to a less favorable outcome under a metric that gives greater weight to large counting errors. This benefit was further limited by CSRNet and P2PNet, which showed higher errors in both metrics. Indeed, real-only CSRNet retained the lowest RMSE across both training settings.

Taken together, these contrasting responses indicate that the proposed subset provides useful additional training data for some counting models, but does not ensure better performance across architectures. Its practical value is supported by the improvements observed for MCNN and APGCC, while the remaining results establish that this benefit depends on the model and the evaluation metric.

\begin{table*}[!t]
\caption{Downstream Crowd-Counting Performance under Different Training Settings}
\label{tab:downstream_counting_performance}
\centering
\footnotesize
\renewcommand{\arraystretch}{1.15}
\setlength{\tabcolsep}{5pt}

\begin{tabularx}{\textwidth}{
>{\raggedright\arraybackslash}p{0.18\textwidth}
>{\centering\arraybackslash}X
>{\centering\arraybackslash}X
>{\centering\arraybackslash}X
>{\centering\arraybackslash}X
}
\toprule

&
\multicolumn{2}{c}{\textbf{Real + CrowdH-Edit-Ann-387}} &
\multicolumn{2}{c}{\textbf{Real Data Only}} \\

\cmidrule(lr){2-3}
\cmidrule(lr){4-5}

\textbf{Models} &
\multicolumn{2}{c}{\textbf{N=694}} &
\multicolumn{2}{c}{\textbf{N=307}} \\

\cmidrule(lr){2-3}
\cmidrule(lr){4-5}

&
\textbf{RMSE} &
\textbf{MAE} &
\textbf{RMSE} &
\textbf{MAE} \\

\midrule

MCNN
& $655.51 \pm 5.62$
& $248.107 \pm 24.01$
& $667.20 \pm 21.61$
& $387.88 \pm 60.77$ \\

CSRNet
& $453.388 \pm 21.99$
& $211.67 \pm 33.49$
& $251.74 \pm 18.77$
& $168.63 \pm 25.98$ \\

DM-Count
& $500.80 \pm 200.02$
& $200.15 \pm 59.80$
& $424.81 \pm 286.92$
& $226.78 \pm 94.65$ \\

P2PNet
& $728.73 \pm 18.47$
& $326.07 \pm 20.54$
& $452.22 \pm 38.90$
& $286.06 \pm 34.46$ \\

APGCC
& $348.77 \pm 9.27$
& $155.53 \pm 27.42$
& $387.60 \pm 19.23$
& $157.84 \pm 9.97$ \\

\bottomrule
\end{tabularx}
\end{table*}

%%%%%%%%%%%%%%%%%%%%%%%%%%%%%%%%%%%%%%%%%%
\section{Limitations}
Although our proposed framework shows promising results, several limitations remain. First, the generated images depend on the diversity and quality of the real reference images. Therefore, locations, viewpoints, lighting conditions, or crowd patterns that are not represented in the reference set may remain underrepresented in the generated data.  Second, although Qwen Image Edit improves reference-scene preservation compared with the LoRA-based baselines, its output may still contain artifacts particularly in dense crowd regions. Third, CrowdH-Edit-Ann-387 remains smaller than large general crowd-counting benchmarks. Future work should expand the annotated subset.

%%%%%%%%%%%%%%%%%%%%%%%%%%%%%%%%%%%%%%%%%%
\section{Conclusion and Future work}
This study presented CrowdH-Edit, a reference-guided framework for generating synthetic Hajj crowd images using Qwen Image Edit within a ComfyUI workflow. The framework addresses limited domain-specific data availability by using real Hajj images and structured editing instructions to vary crowd conditions and illumination while helping to preserve the main scene layout. The cleaned image pool provided the basis for CrowdH-Edit-Ann-387, an annotated synthetic subset for downstream crowd-counting evaluation.
The evaluation of image quality favored Qwen Image Edit over the SD 2.1 + LoRA and SDXL 1.0 + LoRA baselines in semantic alignment, distributional fidelity, and reference-based structural and perceptual similarity. However, the downstream benefits of the Qwen-generated subset varied between the counting models. Adding CrowdH-Edit-Ann-387 reduced MAE for three of the five models and RMSE for two. These findings support the practical usefulness of the proposed subset for some counting models, while showing that its benefits depend on the model and evaluation metric. The framework provides a practical approach to expanding the Hajj crowd imagery, although reference-image diversity, generated artifacts, and the size of the annotated subset remain limitations. Future work will expand the annotated subset, improve high-density sample quality, include more Hajj locations and viewpoints, and evaluate the dataset using additional crowd-counting models and larger real test sets.

% \newpage   % <-- احذف هذا السطر (هو سبب الفراغ الكبير)

% ===== Appendix starts immediately (no blank page) =====
%\clearpage

% ==========================================
% APPENDIX A
% ==========================================

\appendices

\appendices
\section{Instruction Design Details for Qwen Image Edit}

\subsection{Representative Instruction Templates}

The following instruction templates were used for
instruction-conditioned Hajj crowd image editing.

\medskip

\noindent
\textbf{Instruction Template A1:}\\
\textbf{Name:}
\texttt{Kaaba\_Very\_Low\_Crowd\_Early\_Morning\_Aerial}

\par\noindent
First, remove the crowd in this photo until only 10 to 20 people remain, scattered realistically. In the empty space, you must create a seamless, plain white marble floor that perfectly matches the perspective. After that is done, re-light the entire scene for early morning, with the sun at a low angle creating cool, soft light and long, diffused shadows.

\par\noindent
\textbf{Final Check:} Do not add any extra walls or buildings. The people must look photorealistic. The white pilgrim clothes must stay pure white.

\medskip

\noindent
\textbf{Instruction Template A2:}\\
\textbf{Name:}
\texttt{Kaaba\_High\_Crowd\_Midday\_Aerial}

\par\noindent
First, fill the scene with a high-density, photorealistic crowd in a circular Tawaf pattern, ensuring people are scaled correctly. Then, re-light the scene for bright midday with a high sun and short, hard-edged shadows.

\par\noindent
\textbf{Final Check:} Do not alter the existing architecture. The crowd must look real. The white pilgrim clothes must stay brilliant white.

\medskip

\noindent
\textbf{Instruction Template A3:}\\
\textbf{Name:}
\texttt{Jamarat\_Medium\_Crowd\_Midday\_Aerial}

\par\noindent
First, remove about half of the crowd from the bridges. In the empty space, create a seamless, plain light-grey concrete floor that matches the perspective. Then, re-light the scene for bright midday with a high sun and short, hard-edged shadows.

\par\noindent
\textbf{Final Check:} Do not add extra buildings. The people must look real. Keep their white clothes pure white.

\medskip

\noindent
\textbf{Instruction Template A4:}\\
\textbf{Name:}
\texttt{MINA\_Golden\_Hour}

\par\noindent
In this photo of Mina, do not change the tents, the crowd, or the layout at all. Only re-light the entire scene for golden hour, with a low sun bathing the tents in warm, golden light and casting long, soft shadows. The viewpoint should be a top-down aerial shot.

\medskip

\noindent
\textbf{Instruction Template A5:}\\
\textbf{Name:}
\texttt{Safa\_and\_Marwah\_High\_Crowd\_Midday\_Aerial}

\par\noindent
First, fill the walkway with a high-density, photorealistic crowd moving in two directions, ensuring people are scaled correctly. Then, re-light the scene to simulate bright midday with intense, neutral white light from all ceiling fixtures.

\par\noindent
\textbf{Final Check:} Do not alter the existing architecture. Preserve the rocks. The crowd must look real. The white pilgrim clothes must stay brilliant white.

\medskip

\noindent
\textbf{Instruction Template A6:}\\
\textbf{Name:}
\texttt{Mount\_Arafat\_High\_Crowd\_Evening\_Aerial}

\par\noindent
First, fill the plain with a high-density, photorealistic crowd arranged in large, organic clusters, ensuring people are scaled correctly for the landscape. Then, re-light the scene for evening, using only a few sparse, temporary artificial lamps to create pools of light and deep shadows.

\par\noindent
\textbf{Final Check:} Do not alter the existing landscape. The crowd must look real. The white pilgrim clothes must stay bright white.

% Continue with Templates A3--A6 here.

% --- IMPORTANT: do NOT put FloatBarrier/clearpage/newpage here ---
%\vspace{-2mm} % tighten gap between caption and the next text (adjust -1mm to -4mm)

%\FloatBarrier

% If References come after, they will now start AFTER Fig.29
%\FloatBarrier
%\twocolumn

\FloatBarrier

\end{document}